\documentclass[final,5p,times,twocolumn]{elsarticle}

\usepackage{pgfplots}
\usepackage{multicol}
\usepackage[colorlinks=true]{hyperref}

\usepackage{bm}
\usepackage{url}
\usepackage{graphicx} 
\usepackage{mathptmx} 
\usepackage[mathscr]{euscript}  
\usepackage{amsmath} 
\usepackage{amssymb}  
\usepackage{flushend}
\usepackage{siunitx}
\usepackage{xcolor}
\usepackage{adjustbox}
\usepackage{array}
\usepackage{multirow}
\usepackage{booktabs}

\usepackage[inline]{enumitem}
\usepackage{microtype}
\usepackage{amsmath}
\usepackage{threeparttable}
\usepackage{caption}
\usepackage[resetlabels]{multibib}

\usepackage{tikz}
\usepackage{pgfplots}
\usepackage{natbib}

\usepackage{color}

\definecolor{Rbrew}{RGB}{228,26,28}
\definecolor{Gbrew}{RGB}{55,126,184}
\definecolor{Bbrew}{RGB}{77,175,74}

\usepgfplotslibrary{groupplots}
\pgfplotsset{compat=1.13}

\renewcommand{\eqref}[1]{Equation~(\ref{#1})}

\newcolumntype{P}[1]{>{\centering\arraybackslash}p{#1}}

\journal{\href{https://www.journals.elsevier.com/robotics-and-autonomous-systems}{Robotics and Autonomous Systems}}

\makeatletter
\def\ps@pprintTitle{%
   \let\@oddhead\@empty
   \let\@evenhead\@empty
   \def\@oddfoot{\reset@font\hfil\thepage\hfil}
   \let\@evenfoot\@oddfoot
}
\makeatother

\makeatletter
\providecommand{\doi}[1]{%
  \begingroup
    \let\bibinfo\@secondoftwo
    \urlstyle{rm}%
    \href{http://dx.doi.org/#1}{%
      doi:\discretionary{}{}{}%
      \nolinkurl{#1}%
    }%
  \endgroup
}
\makeatother

\begin{document}

\begin{frontmatter}

\title{From Learning to Relearning: A Framework for Diminishing Bias in\\Social Robot Navigation}

\author{Juana Valeria Hurtado\corref{cor1}}

\author{Laura Londo\~no\corref{cor1}}

\author{Abhinav Valada}



\cortext[cor1]{\textit{Equal contribution.}}

\address{Department of Computer Science, University of Freiburg, Freiburg, Germany}

\begin{abstract}
The exponentially increasing advances in robotics and machine learning are facilitating the transition of robots from being confined to controlled industrial spaces to performing novel everyday tasks in domestic and urban environments. In order to make the presence of robots safe as well as comfortable for humans, and to facilitate their acceptance in public environments, they are often equipped with social abilities for navigation and interaction. Socially compliant robot navigation is increasingly being learned from human observations or demonstrations. We argue that these techniques that typically aim to mimic human behavior do not guarantee fair behavior. As a consequence, social navigation models can replicate, promote, and amplify societal unfairness such as discrimination and segregation. In this work, we investigate a framework for diminishing bias in social robot navigation models so that robots are equipped with the capability to plan as well as adapt their paths based on both physical and social demands. Our proposed framework consists of two components: \textit{learning} which incorporates social context into the learning process to account for safety and comfort, and \textit{relearning} to detect and correct potentially harmful outcomes before the onset. We provide both technological and societal analysis using three diverse case studies in different social scenarios of interaction. Moreover, we present ethical implications of deploying robots in social environments and propose potential solutions. Through this study, we highlight the importance and advocate for fairness in human-robot interactions in order to promote more equitable social relationships, roles, and dynamics and consequently positively influence our society.
\end{abstract}

\begin{keyword}
Social Robot Navigation \sep Robot Learning, Fairness-Aware Learning \sep Algorithmic Fairness \sep Ethics
\end{keyword}

\end{frontmatter}


\section{Introduction}
The last decade has brought numerous breakthroughs in the development of autonomous robots which is evident from the manufacturing and service industries. More interesting are the advances that are essential enablers of several innovative applications such as robot-assisted surgery~\citep{tewari2002technique}, transportation~\citep{thrun1995approach}, environmental monitoring~\citep{valada2012intelligent}, planetary exploration~\citep{toupet2020terrain} and disaster relief~\citep{mittal2019vision}.
Novel machine learning algorithms accompanied by the boost in computational capacity and availability of large annotated datasets have primarily fostered the progress in this field. Machine learning and reinforcement learning techniques enable robots to learn complex tasks directly from raw sensory input. 
One such task of navigation has seen tremendous progress over the years. Robots today have the capability to autonomously plan paths to reach a certain location and even make decisions based on the scene dynamics, avoiding collisions with people and objects~\citep{gaydashenko2018comparative, boniardi2016autonomous, jamshidi2019machine, hurtado2020mopt}. 
Advancing robot navigation abilities is crucial for robots to effectively operate in real-world environments.

Robot navigation is a complex task that requires a high degree of autonomy. 
For a robot to successfully navigate the real-world, it is essential to fulfill high accuracy, efficacy, and efficiency requirements. Additionally, it is critical to consider safety standards while developing robots that navigate around humans. To carry out this task, robots are equipped with sensors that allow them to perceive the environment and a path planning system that enables them to compute a feasible route to achieve the navigation goal. So far, mobile robots have been successfully employed in various applications such as material transportation, patrolling, rescue operation, cleaning, guidance, warehouse automation, among others~\citep{nolfi2002synthesis, poudel2013coordinating, bogue2016search, hasan2014path}. This also elucidates that mobile robot applications are moving closer from the industry to everyday tasks in households, offices, and public spaces.
Robot navigation models tailored to solely reach a goal location efficiently are insufficient in these spaces where robots cohabitate with humans. Other complex considerations such as social context, norms, and conventions are essential to ensure that the presence and movements of robots are safe and comfortable.
These additional considerations of sociability play an indispensable role in the acceptance of robots in human spaces. Nevertheless, modeling the social policies that represent humans is a challenging task. To better capture the social behavior of navigation, several learning approaches have been proposed with the goal of directly imitating human navigation or learning from demonstrations~\citep{khambhaita2020viewing, wittrock2010learning, bicchi2015social, silver2010learning}.
With the aim of incorporating social context in learning algorithms, socially-aware robot navigation extends the traditional objective of reaching a certain location to also reflect social behavior in the decision making process~\citep{kretzschmar2016socially}. This can be achieved with learning methodologies based on social and cultural norms. 
These social characteristics can be incorporated into the learning process as social constraints~\citep{khambhaita2020viewing, wittrock2010learning, bicchi2015social} or via imitation and demonstrations~\citep{silver2010learning}. As the role of robots within society is that of a social agent, they should follow social conventions for better acceptability in human environments. Following such conventions will enable them to generate actions that are influenced by respecting personal spaces, perceiving emotions, gestures, and expressions~\citep{luber2012socially, kretzschmar2016socially, ferrer2013social, kruse2013human}. 


However, despite significant advances that enable incorporating social conventions into navigation models, there is still no guarantee that a socially-aware robot will always make fair decisions.
We can extensively observe in other applications of machine learning and Artificial Intelligence (AI), how learning algorithms replicate, promote, amplify injustice, unequal roles in society, and many other societal as well as historical biases. Numerous cases have been identified in face recognition, gender classification, and natural language processing methods~\citep{buolamwini2018gender, garcia2016racist, wang2020visual, lu2020gender, costa2019analysis, benthall2019racial, wilson2019predictive}.
Similar to these cases, learning social behavior from real-world observations will not prevent discrimination. This is of special concern in service and caregiving applications where robots physically interact with humans.

There are multiple social and technical factors that can lead to bias while learning social robot navigation models. First, learning techniques require guidance to optimize the navigation model. Supervised approaches utilize datasets gathered from simulations, controlled experiments, or the real-world. Other approaches, such as imitation learning and reinforcement learning, obtain guidance directly from real experiences. It is important to consider that real-world data can always include bias reflecting unwanted humans behaviors. Additionally, simulations and controlled experiments cannot contain sufficient diverse information about diverse groups of people and their interactions for the robot to learn the large number of potential unfair situations that it can encounter. Therefore, current learning algorithms can significantly replicate, promote, and amplify unfair situations. Besides data-related issues, learning algorithms tend to find certain features that make it easier to optimize for a task and rely on these attributes to learn the function or policy. This can lead to mechanisms that depend on these potential bias inducing features related to a particular characteristic such as race, age, or gender. Another issue encompasses fairness measurements. Thus far, there are no standard fairness definitions or metrics for the optimization of learning-based navigation algorithms or even to detect biased or unfair situations. Furthermore, robots are typically deployed with models that have been pre-trained and do not have the ability to automatically update their parameters or their policy online if they encounter a discrimination scenario.


Recently, several strategies to mitigate unfair outcomes in learning algorithms for tasks such as classification or recognition have been proposed~\citep{agarwal2018reductions, zafar2017fairness, dixon2018measuring, woodworth2017learning}.
Nevertheless, learning fair social navigation models for robotics is substantially lesser studied. Particularly, investigating fairness in mobile robot navigation presents more complex challenges that are not manifested in other data-driven tasks in computer vision and machine learning. In learning-based mobile robot navigation, fairness behavior not only depends on data but also on the future actions of the humans around the robot and other factors of the environment. In this case, it is impractical to anticipate all the possible actions in advance during the development of these models. With these considerations in mind, socially-aware robot navigation, besides learning social skills, should also account for non-discriminatory and fair behavior that makes the interaction safer for diverse groups of people.

In the case of humans, the learning process is not fixed but rather continuous. This allows humans to have both physical and social adaptability. We refer to this adaptive learning from experiences as relearning in this work. We, as humans, not only relearn about the physical world to react to unexpected obstacles in our path, but we also develop adaptability in terms of interaction. This generally prevents us from causing harm to others with our actions and enables us to correct our behavior when we encounter unfair situations. Within this social adaptation, we learn to behave socially and fairly with those with whom we relate to~\citep{mcdonald2008proactive, hutchins2006distributed, goodwin2000action}. The relearning process allows us to reason about what we are experiencing and develop a personality defined by certain moral values, ethical values, beliefs, and ideologies, which in turn influences the way we interact with others~\citep{jarvis2006towards}. Humans decide how to navigate in public spaces while taking both social conventions and ethical aspects into account, such as empathy, solidarity, recognition, respect for people, and recognizing behaviors that lead to discrimination. Accordingly, learning and relearning are important processes for humans to acquire the capabilities that are required for navigating in the environment and cohabitate in society.

Inspired by the learning and relearning processes in humans, we propose a framework for diminishing bias in social robot navigation. Our framework consists of two components. During robot development, we introduce social context based on social norms and skills while learning navigation models so that the robot acquires social conventions. We then incorporate a relearning mechanism that detects systematic bias in control decisions made by the robot during navigation. This enables the robot to update its navigation model when unfair situations are detected during the operation. Our proposed framework facilitates diminishing bias in the behavior of the robot and generates early warnings of discrimination after the deployment. More importantly, it enables the adaptation of the robot's navigation model to new cultural and social conditions that are not considered during training. 

In this work, we describe the motivation and the technical approach for implementing our proposed Learning-Relearning framework for social robot navigation. We then highlight the risks and propose potential solutions that include specific fairness considerations for mobile robots that navigate in social environments.  
Furthermore, we analyze the ethical and societal implications of deploying mobile robots in social environments. To this end, we investigate the behavior of mobile robots in terms of fairness in three specific service and caregiving scenarios with different levels of human-robot interaction. There are other social scenarios where the mobility of the robot directly depends on the human's control action such as autonomous wheelchairs~\citep{johnson2018socially} or robotic guide canes~\citep{ulrich2001guidecane}. Nevertheless, in this work, we only consider scenarios where the robot navigates as an independent machine that interacts with multiple humans in the surrounding environment at different levels of priority. We provide examples that show cases where models that are only based on learning social navigation are insufficient to obtain fair behavior, and we discuss how the relearning mechanism can extend those models to yield fair behavior. Finally, we analyze scenarios in which learning social behavior and accounting for fair behavior play an important role in the real-world.

To the best of our knowledge, this is the first work to investigate the societal implications of bias in learned socially-aware robot navigation models, and the framework that we present is the first to demonstrate a feasible solution for learning fair socially compliant robot navigation models. Even though our work targets socially-aware robot navigation, the framework that we propose can also be extended to other aspects of human-robot interaction, which would benefit from the presented insights. As a result of the social perspective, we provide a comprehensive understanding of fairness in human-robot interactions. This is an important step towards diminishing bias and amplifying healthy social conventions to positively influence the society. With this work, we aim to create awareness that robots should positively impact society and should never cause harm, especially against individuals or groups who have been historically marginalized and who disproportionately suffer the unwanted consequences of algorithmic bias.

In summary, the primary contributions of this paper are:
\begin{itemize}[noitemsep,leftmargin=2\parindent,labelsep=1\parindent,topsep=-2pt]
    \item We introduce a framework for diminishing bias in social robot navigation, consisting of two stages: Learning and Relearning. We present the technical concept and introduce methods that can be used to implement our framework.
    \item We present a societal and technical analysis of the social abilities and bias considerations in learning robot navigation models.\looseness=-1
    \item We present the social implications of socially-aware robot navigation models and provide a set of fairness considerations.
    \item We provide detailed case studies that analyze the impact of bias in different service and caregiving robot applications and discuss mitigation strategies.
\end{itemize}
 
\section{Ethical Aspects and Fairness Implications}

\begin{figure*}
\centering
\begin{tikzpicture}
\pgfplotsset{compat=1.3}
\begin{axis}[width=8cm,
xtick=data,
xticklabel style=
{/pgf/number format/1000 sep=,rotate=60,anchor=east,font=\scriptsize},
minor tick num=2,
ylabel=Papers,
xlabel=Year,
scaled y ticks = false,
xmajorgrids=true,
ymajorgrids=true,
yminorgrids=true,
grid style=dashed,
width=15cm,height=8cm,
legend style={at={(1,-0.27)}},
every axis plot/.append style={ultra thick},
legend style={legend columns=-1}
]
\addplot[color=Bbrew,
    mark=square,
    domain=2007:2020, 
    ]
coordinates {
    (2011,10872)
    (2012,14843)
    (2013,17826)
    (2014,22833)
    (2015,15061)
    (2016,15669)
    (2017,19307)
    (2018,23490)
    (2019,24767)
    (2020,25379)};
\addlegendentry{Robot Navigation}

\addplot[color=Gbrew,
    mark=square,
    domain=2007:2020, 
    ]
coordinates {
    (2011,3420)
    (2012,6778)
    (2013,6020)
    (2014,4259)
    (2015,4788)
    (2016,4538)
    (2017,6887)
    (2018,8969)
    (2019,9040)
    (2020,10772)};
\addlegendentry{Social Robot Navigation}

\addplot[color=Rbrew,
    mark=square,
    domain=2007:2020, 
    ]
coordinates {
    (2011,1939)
    (2012,4736)
    (2013,3183)
    (2014,2293)
    (2015,2234)
    (2016,2395)
    (2017,3077)
    (2018,4863)
    (2019,4769)
    (2020,6021)};
\addlegendentry{Fair Robot Navigation}
\end{axis}

\end{tikzpicture}
\caption{Comparison of the number of publications on Robot Navigation (blue), Social Robot Navigation (red), and Fair Robot Navigation (green) from 2011 to 2020. Although the rate at which fairness is being considered in robot navigation methods is increasing, there is a growing gap with the number of works that address robot navigation each year.}
\label{fig:plot}
\end{figure*}
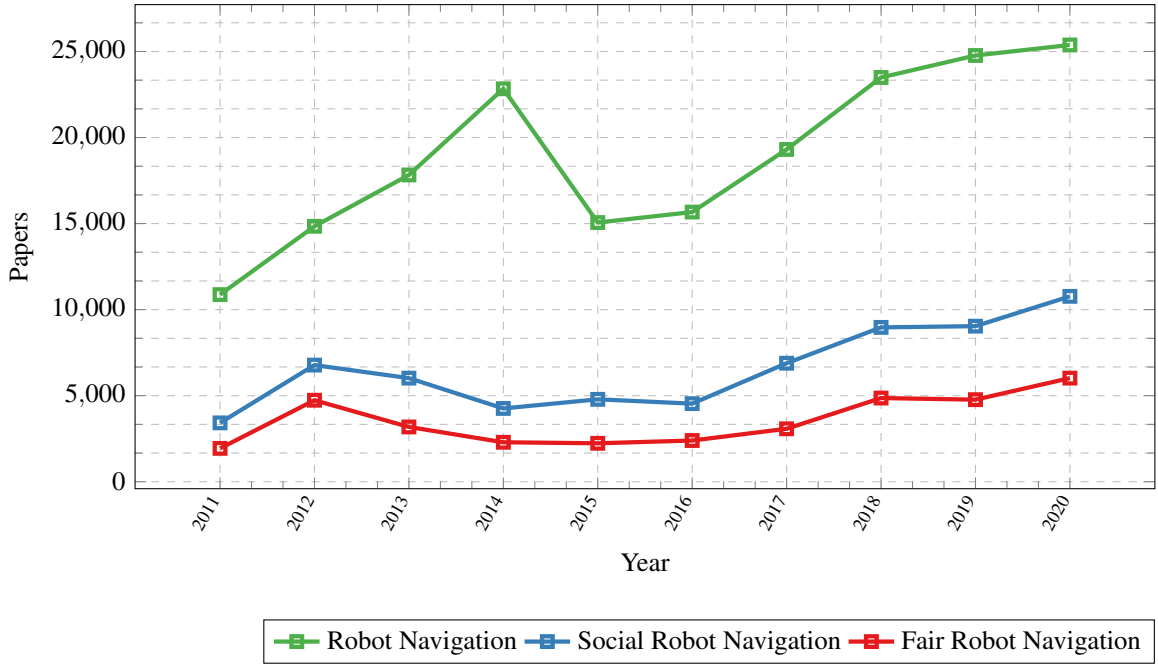

The growing impact that AI and robotics have in the daily lives of people has led to the increase in ethical discussions about current machine learning algorithms and how to handle new research towards an equal and positive impact of technology for diverse groups of people. Consequently, recent works in both social sciences and machine learning have highlighted the challenges in socio-cultural structures that are reflected and amplified by learning algorithms. As a result, many guidelines from the technical~\citep{hagendorff2020ethical,piano2020ethical,silberg2019notes,cath2018governing} and social perspectives~\citep{liu2017responsible,birhane2019algorithmic,verbeek2008morality} have been presented. These guidelines~\citep{vayena2018machine, piano2020ethical, hagendorff2020ethics} are aimed towards mitigating the adverse effects and advocating for ethical principles such as fairness, trust, privacy, liability, data management, transparency, equality, justice, truth, and welfare. Similar efforts have been made by the European Robotics Research Network (Euronet) in the Euronet Roboethics Atelier project in 2005, and the British Standards Institute which published the World's First Standard on Ethical Guidelines in 2016~\citep{torresen2018review}. Moreover, some works in robotics~\citep{lin2012robot, anderson2010robot, boden2017principles, bsi2016bs} have also investigated the importance of addressing ethical issues for safe and responsible development.

These ethical guidelines~\citep{reed2016responsibility, johnson2019artificial, arrieta2020explainable, goodman2017european} share the value of robots effectively and safely assisting people, and under no circumstance cause harm or endanger their physical integrity~\citep{de2008atlas, vandemeulebroucke2020ethics, riek2014code}. The impact of human-robot interactions has also been studied to a lesser extent in mobile robotics, e.g., providing recommendations on road safety, privacy, fairness, explainability, and responsibility~\citep{bonnefon2020ethics}, or studying fairness in path planning algorithms of robots during emergency situations~\citep{brandao2020fair}. Similarly, such ethical discussions should be contrived while developing socially-aware robot navigation models. As shown in Figure~\ref{fig:plot}, although the number of publications that consider fairness in robot navigation is slowly increasing, it is still over five-times lesser than the overall number of publications that address robot navigation.
In this section, we present a series of ethical aspects and social implications that can arise from bias in socially aware-robot navigation algorithms. Additionally, we analyze the impact that these social navigation algorithms can have on human environments.

\subsection{Fairness Implications}
\label{subse:fairness_impl}

The cultural and social knowledge in humans is transferred from generations as a cumulative inheritance that allows each member of the society to incorporate moral, political, economic, and social structures that not only have a positive but also a negative value~\citep{castro2004evolution}. These inheritance conditions have perpetuated historical discrimination against individuals and groups of people. The data collected in machine learning and AI come from these historical inheritance structures; consequently, social-historical discrimination can also be reflected or even amplified by learning algorithms. In recent years, several unexpected outcomes have been observed in learning algorithms that have caused discrimination and prejudice in society. Numerous examples demonstrate how social prejudices are reflected in machine learning algorithms~\citep{garcia2016racist, wang2020visual}. One clear example that was observed in natural language processing was the racial and gender biases while learning language from text~\citep{lu2020gender,costa2019analysis}. Another recent example is the automated risk assessments used by U.S. judges to determine bail and sentencing limits. It was shown that it can generate incorrect conclusions, resulting in large cumulative effects on certain groups, such as longer prison sentences or higher bails imposed on darker-skinned users~\citep{benthall2019racial}. Moreover, another study shows how biased algorithms affect the performance of vision-based object detectors employed in autonomous vehicles. Their work demonstrates that pedestrians with dark-skinned tones presented higher recognition errors~\citep{wilson2019predictive}. There have also been numerous cases of algorithmic bias that have been observed in algorithms used in healthcare. For example, algorithms trained with gender-imbalanced data have shown higher error at reading chest x-rays for an underrepresented gender~\citep{amit2020}.

The numerous cases of discrimination observed in learning algorithms employed in various applications are a source of concern for robotics. In the case of robots that employ learning algorithms to effectively interact, navigate and assist people, it is essential to foresee possible unfair situations. Specifically, as a result of learning socially-aware robot navigation strategies, these trained models can enhance the social impact in terms of human acceptance of mobile robots, daily use, comfort, security, protection, and cooperation~\citep{thrun2000interaction}. Providing robots with a more natural navigation ability also increases their usability. 
Although incorporating social navigation models in robots improves their usability, comfort, and safety in human spaces, social abilities by themselves do not ensure fair robot decisions, especially while using learning algorithms to imitate or follow human conventions and behaviors. In human social interactions, a series of direct and indirect discrimination behaviors and decisions are often present~\citep{yu2019direct,forshaw2008direct,zhang2016causal}. 
Using learning algorithms can negatively affect society, individuals, or groups if unwanted social behavior is replicated and reflected in the actions of the robot. Therefore, this highlights the need to implement fairness considerations and measures. The ability of an agent to dynamically make fair decisions among different people is a fundamental basis for trust in human-robot interaction~\citep{otting2017criteria, claure2019reinforcement}. If robots, after their deployment, present an unfair behavior, it will continue to perpetuate discriminatory structures that will be reflected in the way that people are assisted. Moreover, this will cause serious consequences such as a large population not being benefited by the robots and being reticent to use them. These factors suggest that the robot would only be beneficial for certain groups of people, which would continue to reinforce large social inequalities. Robots should influence society in a positive way by promoting healthier relationships, roles, and dynamics after their deployment in different places with diverse people. This requires the creation of a more reflective, equitable, and inclusive learning methods accompanied by extensive studies from the social perspective.

\subsection{Fairness Measures}
\label{sec:fairnessmeasures}


Fairness is a complex ethical principle that relates to avoiding any form of systematic discrimination against certain individuals or groups of individuals based on the use of particular attributes such as race, sexual orientation, gender, disability, socioeconomic and sociodemographic position~\citep{silberg2019notes}. However, the definition of fairness tends to be dynamic, mobile, and contingent; therefore, it should be analyzed from a reflective and ethical perspective. Moreover, fairness highly depends on the context, location, and culture, among other factors. Consequently, defining an accurate fairness measure could be a complex task. With efforts in this direction, bias has been used to represent fairness either in human environments or in technological developments~\citep{lee2018detecting, nelson2019bias, fuchs2018dangers, howard2017addressing}.

For its part, solutions to algorithmic bias that perpetuate social and historical discrimination against vulnerable and disadvantaged individuals or groups of people tend to be technical rather than moral and ethical~\citep{birhane2019algorithmic}. Technological solutions to biased decisions making are essential but not solely sufficient. Instead, technical solutions should be accompanied by factors such as diversity, inclusion, and participation of underrepresented groups during the development of navigation models. Although there is no standard definition of fairness in machine learning and AI, some works state that a prediction is fair when it is not discriminating or when there is no bias~\citep{birhane2019algorithmic,chouldechova2018frontiers,binns2018fairness}. However, there are two types of biases, positive and negative. Positive bias frequently promotes social good and avoids prejudice through awareness and respect for human differences. Therefore, not all biased outputs are necessarily undesirable, and eliminating them can cause unintended outcomes for certain people. For example, consider an algorithm that is used in a bank to perform a credit study of the people who apply for a loan. If the algorithm is trained to guarantee that all the people will have credit, this may be a disadvantage in the long run for those who cannot pay back later. While the algorithm is being equal in this case, it is unfair in the long term as it negatively affects the low-income people~\citep{silberg2019notes}.

In socially-aware robot navigation fairness measurements are yet to be studied. As robots interact and assist different groups of people in different settings, creating a unified definition or metric is impractical due to the complex and diverse cases that robots can encounter after deployment. Accordingly, in order to tackle unfairness, we present a series of fairness considerations for socially-aware robot navigation:
\begin{enumerate}[noitemsep,label=(\roman*),leftmargin=2\parindent,labelsep=1\parindent,topsep=-2pt]
    \item \textbf{Value Alignment} refers to the alignment of human values in decision making during navigation. These values include respect, inclusion, empathy, solidarity, recognition, and non-discrimination. In socially-aware robot navigation, it is reflected in cases when the decision-making of the robot reproduces and increases the welfare of vulnerable populations. For example, prioritizing to assist and serve people with physical disabilities in crowded environments.
    \item \textbf{Bias Evaluation} is related to the evaluation of bias in decisions making during navigation. Bias can be considered acceptable if there is adequate reasoning or unacceptable if the bias replicates, promotes, or amplifies discrimination. For example, when robots navigate with a different speed around young people who are faster than around older adults, it is usually accepted because they have important physical differences. Nevertheless, if such decisions are made based on racial differences, it can be considered unacceptable, given that there are no fair reasons for this difference. With this fairness consideration, when biases are presented in navigation models, it can only be accepted if there are fair reasons for doing so.
    \item \textbf{Deterrence} is expressed in preventing and mitigating unwanted bias as well as discrimination during navigation. Since the notion of deterrence is dynamic and can vary depending on the social context, robots should be sensitive to cultures by adapting to people, customs, and their surroundings. 
    \item \textbf{Non-Maleficence} signifies that the decisions of a robot can never produce damage to people. The damage is primarily interpreted as bodily harm, collisions, interruptions, delay, and obtrusion. However, damage can also refer to the negative effects caused by discrimination, segregation, and bias. For example, if a caregiving robot in a hospital becomes an obstacle to the medical personnel responding to an emergency due to biased decisions, then it would be violating this property. 
    \item \textbf{Shared Benefit} refers to providing equal benefits to diverse people in all scenarios. If a robot is specifically designed for and only tested in a particular geographical area, tailored to the characteristics and behaviors of the people in that region, it can lead to unwanted bias when it is deployed in a new region which may have completely different characteristics. Therefore, the benefits that the robot provides should not be targeted towards people with specific characteristics in a determined geographical area, but should rather be equally beneficial to all users. In this case, adaptability is an important attribute for robots to achieve shared benefit so that the autonomy of the robot is flexible to adapt to characteristics of specific users in the social environment where it is deployed.
\end{enumerate}

\subsection{Responsible Innovation}

Research in technology studies suggests that the conceptions of responsibility should build upon the understanding that science and technology are not only technically but also socially and politically constituted~\citep{winner1978autonomous, grunwald2011responsible}. Responsible Innovation (RI) was introduced as a concept to address the impact of research and innovation in technology from an ethical and fair perspective. RI states that the technology should be anticipatory, so it should have a foresight guide that provides alternative options for responsible development~\citep{brandao2020fair, stilgoe2013developing}, and it should account for social, ethical, and environmental issues. Based on RI principles, the framework that we present in this paper aims to identify biased behavior during navigation and promotes fair decision making through the learning and relearning process to enable flexible and adaptive service. RI articulates and integrates four factors: 
\begin{enumerate*}[label=(\roman*)]
\item anticipation of damages,
\item reflection  from an ethical perspective,
\item protection of sensitive human characteristics, such as age, gender, and race, and 
\item responsiveness~\citep{stilgoe2013developing}.
\end{enumerate*}

With the aforementioned RI factors, responsible robotics aims to ensure that responsible practices are carefully accounted for within each stage of design, development, and deployment. Correspondingly, robot navigation models should address the ethical and legal considerations at the time of development. Given that these considerations are constantly changing depending on the social or cultural factors, these models should be updated accordingly.


\section{Learning - Relearning Framework for Socially-Aware Robot Navigation}

\begin{figure*}
\centering
\footnotesize
\includegraphics[width=\textwidth]{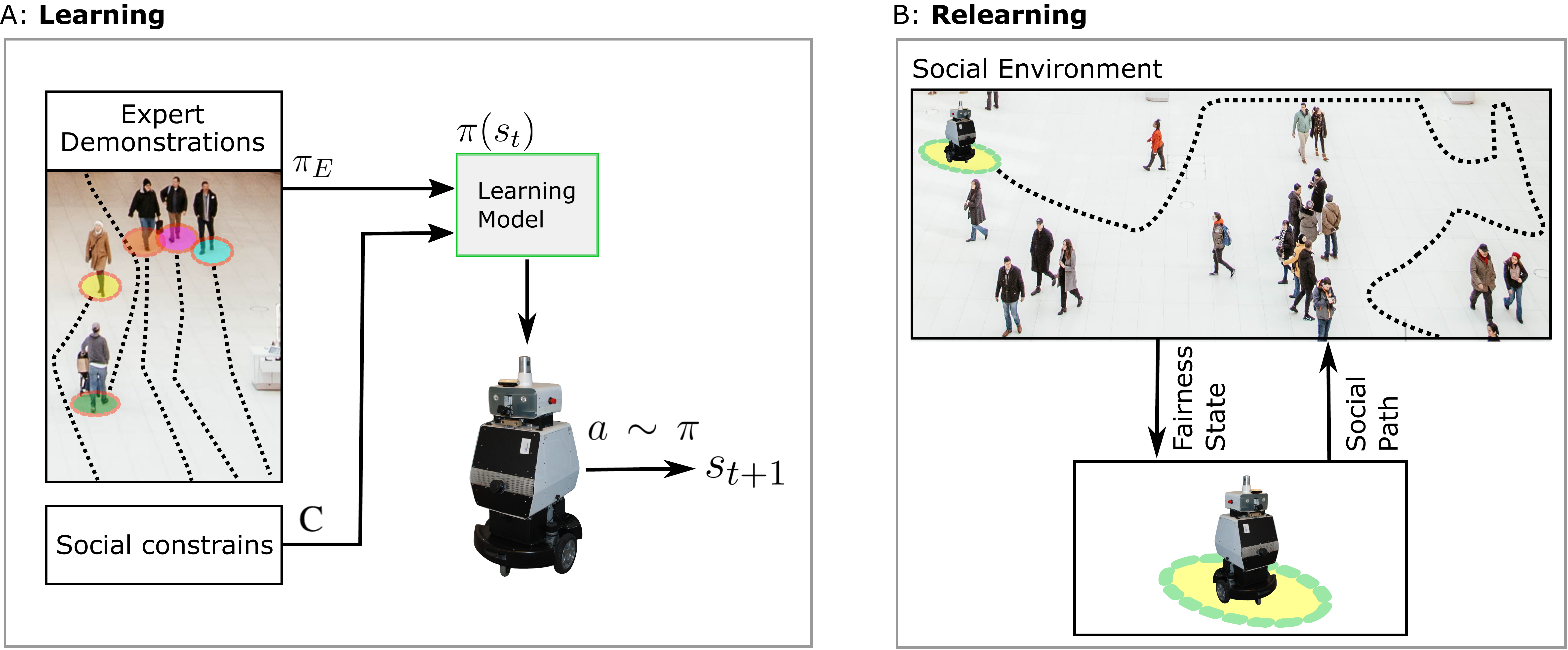}
\caption{Illustration of our proposed Learning-Relearning framework for diminishing bias in social robot navigation. Our proposed framework consists of two components: learning (A) and relearning (B). By including the social context in the learning process, we aim to account for safety and comfort. The social context is presented as the social skills demonstrated by experts and social norms as constraints. Moreover, we aim to detect potentially harmful outcomes before the onset using the relearning mechanism. After detecting unfair effects, the navigation model should be automatically updated to account for fairness.}
\label{fig:scheme}
\end{figure*}

The goal of our proposed framework is to develop learning models for robot navigation that yield social and fair behavior. To this end, we define two different stages: learning and relearning. In the first stage, we incorporate social context into learning navigation strategies so that robots can navigate in a socially compliant manner. While, in the second stage, we aim to diminish any bias in the planned paths with the learned navigation model. In this section, we first introduce socially-aware robot navigation. We then describe our proposed framework and present the technical approach that can be used for the implementation. Figure~\ref{fig:scheme} shows the different stages of our framework. In the learning phase, we learn a navigation policy based on imitation learning with additional social constraints. Whereas, in the relearning phase, we analyze the outputs of the network online and provide the model with updates to reach the navigation target while accounting for and deterring bias to ensure fairness. Science and technology, from the RI perspective, have the ability to provide significant benefit through well-established methodologies that reflect responsibility and ethical principles. This framework tailored exploits the learning and relearning process as a methodology to achieve responsible robot navigation.   

\subsection{Socially-Aware Robot Navigation}
\label{subsec:social_rob_nav}

One of the widely studied requirements for mobile robots to operate in human spaces is the ability to navigate according to social norms and socially compliant behavior. The social navigation models that are employed in robots play an important role in the effect that these automated machines have on society and the perception as well as confidence that humans will have of them. 
In the case of humans, we develop the ability to navigate while considering numerous variables representing the environment, such as the objects, people, and dynamics of the agents in it. This ability, known as sociability, from an anthropological point of view, is the human capacity to cooperate and engage in joint behavior with others~\citep{simmel1949sociology}. Further, sociability allows us to navigate while avoiding situations that make us uncomfortable or put us or others in danger. 

Different social norms have been developed to provide information about the appropriate behavior, especially in public spaces. Social norms are standards of conduct based on widely shared beliefs of how people should behave in a given situation~\citep{fehr2004social}. Some of the social norms for navigation are not invading the personal space of people, passing on the right, maintaining a safe velocity, not blocking people's path, approaching people from the front, among others~\citep{kirby2010social}. Besides social norms, different studies such as proxemics~\citep{hall1968proxemics}, kinesics~\citep{birdwhistell2010kinesics} and gaze~\citep{argyle1976gaze} also provide cues to determine the appropriate manner to approach a person, navigate around, and coordinate in public spaces. Specifically, proxemics is the study of the perception and organization of the personal and interpersonal space. It is associated with the manner of how humans manage their surrounding space when they walk in public environments and how their comfort can be affected by the movement of other pedestrians~\citep{rios2015proxemics}. Kinesics is related to the actions of the body and positions~\citep{birdwhistell1952introduction}, and gaze refers to the eye movements and directions during visual interaction~\citep{harrigan2005proxemics}. These studies highlight social skills such as reading emotions and the prediction of the intentions of people. The combination of both social norms and social skills can be considered a determinant of sociability. 
The aforementioned studies and norms are some of the increasingly used factors in learning social robot navigation models. It is long believed that equipping robots with these social skills and social norms will enable them to react socially as humans do.

For instance, we can anticipate that cleaning robots~\citep{fiorini2000cleaning} that are primarily used in houses will be widely used in public spaces in the coming years. Currently, these robots do not conform to any social norms during navigation. Confined to private locations and users who know the device, manufacturers have not made it a priority to include social skills such as predicting the intention of people and avoiding crashing into them. Nevertheless, sociability is an important skill to deploy cleaning robots in crowded public spaces. In this case, robots must take into account aspects such as the space that they occupy and the personal space of the people around to determine how close to navigate around them or predict where humans will move so that they do not interfere with their paths. These skills will allow robots to plan a safe route so that their presence is not disturbing, surprising, or scaring the people that share the same space. While planning routes, robots should use social norms such as not invading the personal space and maintaining a safe speed. Both the use of social skills and social norms change depending on the type of robot and the context in which it is used. We present further discussions of this example in Section~\ref{sec:cleaning}.

Socially-aware robot navigation methods 
can primarily be categorized into two groups. The first category is model-based and consists of handcrafted models that use mathematical formulations to combine a set of effects to determine the dynamics of pedestrians, such as reaching the destination, the influence of other pedestrians, keeping a certain distance to another person, or the maximal acceptable speed. \cite{helbing1995social} introduced the notion that social forces determine human motion and propose the Social Force Model (SFM) to represent pedestrian dynamics. To navigate in a manner similar to humans, this formulation was later used to provide robots with pedestrian-like behavior for human-robot social interaction~\citep{ferrer2017robot}. However, SFM requires us to cautiously define and tune the parameters for each specific scenario, which makes it impractical to scale to complex tasks and environments~\citep{tai2018socially}. The second category consists of learning-based methodologies that use some form of guidance or demonstrations containing the policies that link observations to the corresponding actions. We further discuss learning-based methods in the following section. 

\subsection{Learning}
\label{subsec:learning}

The rapid progress in machine learning in the past years and the growth of computing power have enhanced the learning capabilities of autonomous mobile robots. Currently, these learning-based methodologies play an essential role in the development of complex navigation models. These models are primarily trained to achieve the best navigation performance under some given metrics during the learning process. For this purpose, different guidance techniques have gained interest in robot navigation works. The first of which is supervision from labeled data, which uses either data gathered from the real-world or simulations and the corresponding annotations. The data and annotations are then employed to optimize the model so that the output predictions are as close as possible to the labels. Supervised navigation methods can be used directly by learning the mapping from the states in recorded trajectories that contain social policies to their corresponding labels or by learning reactive policies that imitate a planning algorithm~\citep{groshev2017learning}.

Another extensively explored learning technique is Reinforcement Learning (RL), in which an agent explores the state and actions by itself while a reward function is used to punish or encourage the decisions to obtain an optimal model. RL techniques can be used to provide a robot with navigation paths that maximize rewards in terms of human safety or comfort~\citep{chen2017socially}. Moreover, Inverse Reinforcement Learning (IRL) is a technique that has been widely used to capture the navigation behavior of pedestrians. Contrary to supervised learning, IRL is able to recover a cost function that explains an observed behavior~\citep{kuderer2013teaching}. The IRL technique proposed by~\cite{hamandi2019deepmotion} trains the social navigation model by learning the navigation policy directly from human navigated paths in order to generate actions that conform to human-like trajectories. 
To include the social context in the learning process, these models aim to clone the navigation behavior of humans. Subsequently, robots are then equipped with these models for socially-compliant navigation. 

Specifically, to clone an expert behavior in the RL framework, consider that an agent in an environment reaches a state $s_{t+1}$ after executing an action $a_t \sim \pi$ that follows a policy $\pi$. At each transition state, the agent obtains a reward $r_t$ presented as a scalar. The goal is for the agent to adjust the policy $\pi$ to maximize the expected long-term rewards that it can receive. Q-learning~\citep{watkins1992q} is an approach that enables us to find an optimal policy based on the state transition set. The $Q$-function represents the value of an action $a_t$ and following a policy $\pi$ as

\begin{equation}
    Q_{\pi} (s_t, a_t) = \textbf{E}[R(s_t)|s_t,a_t],    
\end{equation}

where 
$R$ is the expected long term reward defined as $R=\sum_{t=0}^{\infty} \gamma^t r_t$, being $\gamma \in [0,1]$ the discount-rate. Given the state $s_t$ and action $a_t$ the Q-function indicates the expected discounted accumulative reward. Using the Q-function, we can estimate an optimal policy $\pi$, which maximizes the expected return. Particularly, no reward function is given in the IRL framework. Therefore, it is inferred from observed trajectories collected by the expert policy $\pi_E$ to mimic the observed behavior.

There are numerous works using RL and IRL that generate human-like navigation behavior in controlled conditions. However, we can more elaborately define how we as humans navigate the environment, using a combination of both social skills and social norms as described in Section~\ref{subsec:social_rob_nav}. Social norms can vary with respect to the context, location, and culture. Extending the social skills of the robot by including social norms is important for social domain adaptation. The social norms that a domestic robot should consider while navigating are substantially different from those that a mobile robot in a hospital should conform to. For example, in order for the robot to navigate in a socially compliant manner in a hospital, it is essential for it to identify emergency situations, understand the priority for interaction, and have fast reaction times, so that the robot can never interfere with the paths of hospital staff and cause accidents or delay the treatment of patients. Given that the context and priorities differ, the reaction also accordingly changes. We explore these cases in the case study that we describe in Section~\ref{sec:case_study}.

Recently, a deep inverse Q-learning with constraints technique~\citep{kalweit2020deep} was introduced. This work presents one such model that allows for the combination of imitating human behavior and additional constraints. This is a novel model-free IRL approach that extends learning by imitation with constraints such as safety or keeping to the right. Using the previous definition of Constrained Q-learning~\citep{kalweit2020interpretable}, it includes a group of constraints C that shapes the possible actions in each state. Besides the Q-function in Inverse Q-learning, it also estimates a constrained Q-function $Q_C$ for which the policy is extracted after Q-learning, considering only the action-values of the actions that satisfy the required constraint. This approach shows promising potential for considering relevant social factors while learning socially-aware robot navigation policies, especially by adding diverse constraints that represent current norms in order to yield socially intelligent and unbiased robot behavior. 

\subsection{Fairness Considerations}
\label{subsec:Tech_Fair}

As with most learning approaches, the method described in Section~\ref{subsec:learning} requires a large number of training examples so that the model learns to yield the desired output. Therefore, it is essential to use either data gathered from the real-world, simulations, or control experiments. With the collected data, developers aim to present representative examples of real-world scenarios or guidance of the desired social behavior during navigation. However, these data collection processes can themselves reproduce biases, and as a consequence, it raises a series of critical concerns. In the specific case of learning socially-aware robot navigation from real-world data, robots can reproduce biased behaviors implicit in human-human interaction. On the other hand, the amount of training data that can be obtained from simulations and control experiments is very limited since only a handful of situations are taken into account. Most data collection processes that do not encompass a balanced set of every possible real-world scenario present a risk for robots trained on them as this could lead to navigation with biased behavior. \textit{These circumstances are considered as bias in the data}. Accurate generalization of scenarios that highly deviate from the training data is an extremely difficult task. To address this factor, recent methods have been proposed to filter data that is used to train the models. For instance, \cite{hagendorff2020ethical} presents a selection process for training data that improves the data quality in terms of ethical assessments of behavior and influences the training of the model. Nevertheless, methods to reduce bias in the data that is used for learning robot navigation models still remain unstudied.

Apart from the problems in dataset collection, there is still a lack of a deeper understanding of the underlying principles and limitations of modern learning algorithms. Especially, a phenomenon known as shortcut learning shows how neural networks learn more straightforward predictors that are not necessarily related to the main task or objective~\citep{geirhos2020shortcut}. A typical example of this phenomenon can be seen in the hiring tool developed by Amazon, which predicts strong candidates based on their curriculum. This tool was later found to be biased towards providing advantages for male applicants. Their model, which was trained on historical human decisions that were made during the hiring process, identified that gender was an important feature for prediction~\citep{dastin2018amazon}. \cite{geirhos2020shortcut} analyses the dependency of outputs to strong predictive attributes found by the model during training.

Data-driven models can contain abstract representations of the data and situations that lead to the prediction. Therefore, it is typically challenging to explain the decisions made by a learned model. To facilitate the fairness analysis, we present an approach that is not solely data-driven, and instead, it implicitly incorporates human interpretations of social dynamics using a model that includes high-level and explainable human notions about social conventions, relationships, and interactions to guide a mobile robot. The purpose of analyzing this approach is to demonstrate that biased behaviors can also be learned from biased demonstrations or observations. We analyze the approach proposed by \cite{patompak2019learning} to predict personalized proxemics areas that correspond to the characteristics of individual people. This approach generates personalized comfort zones of a specific size and shape by associating the personal area with the activity that a person performs or characteristics of the person. Using these social descriptions, it estimates the proxemic zone that better matches each pedestrian in the scene. Consequently, the approach relies on personalized boundary delineation of two different areas: one area where the human-robot interaction can occur, and another area that is private, which the robot should avoid navigating through. The approach consists of three parts: human-social mode, learning the fuzzy social model, and a path planner. The human social model utilizes proxemics theory and aims to reflect the pedestrians' social factors in the scene. The social factors that are considered include gender, relative distance, and relationship degree. Using these factors, the approach yields the parameters that determine the private zone of comfort for each person in the scene based on the fuzzy logic system. For each social factor that is considered, the approach defines a membership function as follows:

A binary function depending on the gender of the pedestrian, which is given by
\begin{equation}
MF_{gender} =
    \begin{cases}
      0, & \text{if}\ \text{gender is \textit{Male}} \\
      1, & \text{if}\ \text{gender is \textit{Female}},
    \end{cases}
\end{equation}
a sigmoid function with relative distance input $r_r$, distribution steepness $a_r$, and inflection point $c_r$ describing \textit{near} or \textit{far} distance defined as
\begin{equation}
MF_{distance} = \frac{1}{1 + \exp(- a_r \times (r_r - c_r))},
\end{equation}
and three Gaussian functions representing the degree of relationship (DR) as familiar (\textit{Fam}), acquaintance (\textit{Acq}), and stranger(\textit{Str}), which is given by
\begin{equation}
MF_{relationship} =
    \begin{cases}
      \mathcal{N}(\mu_{\text{Fam}}, s{^2}_{\text{Fam}}), & \text{if DR is}\ \textit{Fam} \\
      \mathcal{N}(\mu_{\text{Acq}}, s{^2}_{\text{Acq}}), & \text{if DR is}\ \textit{Acq} \\
      \mathcal{N}(\mu_{\text{Str}}, s{^2}_{\text{Str}}), & \text{if DR is}\ \textit{Str}.
    \end{cases}
\label{eq:MFrelationship}
\end{equation}

Subsequently, the fuzzy social model is learned from human feedback using an RL approach. The defined membership functions of the social factors can be learned to yield an improved personal area for each pedestrian. This is performed by adjusting the relationship degree in the MF Equation~\ref{eq:MFrelationship} to update the social map. The reward of the RL model is then obtained from human-robot interaction by means of the emotion or feeling of each corresponding person. Therefore, the approach sets the focus on the degree of the relationship to be learned. Finally, the approach selects a path planner that chooses an optimal navigation path in the social cost map. The consequently designed social interaction area using fuzzy rules presents the output of the model as two separated personal areas: far personal area (FPA) and near personal area (NPA). As part of the rules presented, it is clear that for the input gender female, the near personal area is never an option. Taking into account that the reinforcement learning algorithm updates the model based on the $MF_{relationship}$, the resulting navigation policy would never allow for human-robot interaction close to women. \textit{This presents a critical bias of the model due to the inclusion of social dynamics.} This is an example where bias appears due to an explicit constrain in the learning algorithm. Not only gender but other factors that may potentially lead to bias as well as other implicit or explicit biases can appear by learning from real-world data. We discuss this technical bias of the aforementioned navigation model with implications and analysis from the social perspective in Section~\ref{sec:case_study}.

Learning robot navigation policies and models that are unbiased requires analyzing how the input is given, how the data is measured, how the data is labeled, what it means for models to be trained on them, what parameters are used, and how social navigation models are evaluated. If models aim to reflect the features of society, we need to question what behaviors should be replicated and promoted. For example, \cite{kivrak2020social} explicitly exclude women in the real-world experiments of their social navigation framework for assistive robots around humans. Their model that aims to yield human-friendly routes was only tested in a corridor where women were excluded based on previous analysis~\citep{jones2006differences}, which affirms gender differences in spatial problem solving. \textit{This represents bias in the evaluation} where the social model of navigation is validated only for a privileged group and can lead to underperformance to the unconsidered after the deployment. This has also been seen before in medical datasets or experiments where women were excluded citing differences in hormonal cycles, which leads to the medicines or medical procedures causing higher side effects for women compared to men. The consequences of these biased experiments or trials have been extensively discussed, which had lead to the inclusion of women in all medical trials~\citep{soderstrom2001researchers}.

The technical bias analysis presented in this section shows cases where the high-level representation of social interaction replicates unequal roles and dynamics that already exist in human interaction. It is a significantly larger risk in the case of learning models for social navigation from demonstrations where the assumption is that the best way to teach a robot to navigate is to enable it to learn directly by observing humans.

\subsection{Relearning}

While learning socially-aware robot navigation models, social biases can be introduced that replicate and even augment the unfair societal dynamics. Most existing socially-aware robot navigation techniques aim to learn social navigation behavior by imitating human navigation. Consequently, it essential to deter biases during the deployment of robots equipped with such models. In this section, we present a mechanism to first detect when the navigation model makes biased decisions, especially against certain groups of people. Subsequently, we use this mechanism to update the model towards yielding more equitable social navigation policies.

There are many situations in the real-world where unequal decisions are desired, such as adapting the speed of the robot near older adults. In this work, we only analyze situations where there is no justifiable reason to yield different actions while interacting with different groups of people. In this case, an unfair or discriminatory system will offer an advantage to a certain group of users or unfavorable interaction to some other groups. Unfair behavior in robot navigation directly affects how users interact with the system. For a mobile robot to amend a discrimination behavior, it is necessary first to detect or measure the biased behavior. An advantage in the case of robots is that the decisions and actions after deployment can be used to measure the degree of biased decisions, for instance, concerning protected characteristics, such as age, gender, and race. Whereas, in the case of bias in deep learning models this task would be significantly harder. For instance, the Microsoft AI Twitter chatbot Tay which learned by interacting with users and presented gender-biased as well as racially offensive tweets \citep{perez2016microsoft}. In this case, it would be necessary to additionally measure the features behind the posted tweets. Given that most robots are designed to move in the world, this characteristic comes for free in terms of the navigation actions that were made based on distance, speed, among other control variables as well as perception, accuracy, and uncertainty.

The robot can gather a dataset or a log by storing its own experiences and their corresponding actions even after deployment. Subsequently, the first step is to detect bias in the social navigation decisions of the robot. Bias identification is related to detecting disproportionate prejudice or favoritism towards some individuals or groups over others. For example, the paths planned by the robot produce a negative effect more frequently for specific groups of people than they do for another, such as discomfort, lack of interaction, or avoidance. Other situations are related to a disproportionate rate of a favorable or higher quality of attributes prediction for certain groups. This situation can present itself due to a lack of representation and diversity in the data or scenarios that were used in the learning stage. As a result, it can lead to unpredictable or no interaction with individuals of these groups. 

One such method to detect if the navigation model exhibits outcomes that differ across subgroups is using clustering. Clustering is the technique for grouping data such that the elements of the same group are assigned closed together, forming assemblies called clusters. Clustering is a well-studied technique that is highly used in unsupervised or exploratory data analytics. Consider that the dataset collected while the robot was navigating contains all the decisions that were taken, as well as the sensor data and the actions of other agents that these decisions were based on. Additionally, other navigation and perception attributes can be considered, such as the relative distance of the pedestrians to the robot, collisions, person identification confidence, and intention prediction, as well as additional information such as rules that were violated and accidents that were caused. The accumulation of actions the robot outputs corresponds to the navigation feature set to be clustered. The resulting clusters can later be correlated to potential protected characteristics.

Having a learned policy $\pi$ for socially-aware robot navigation, we define $V = \{v_1,v_2,...,v_i\}$ as the set of navigation data that correspond to the experiences that the robot continuously accumulates through certain time steps. Different clustering algorithms can be used depending on the attributes of the selected navigation features (for instance if their nature is categorical or numerical). One promising clustering algorithm is the method proposed in \cite{aljalbout2018clustering} which consists of a fully convolutional autoencoder trained with two losses, one for reconstruction and the other for cluster hardening. The result of the clustering process is a collection of assemblies $A = {A_1,A_2,...,A_K}$ consisting of navigation feature combinations. Each $A_k$ represents the navigation experience that is similar enough to be considered as a cluster of the entire set $V$. The number of clusters $K$ and the size of each cluster $A_k$ are hyperparameters that can be explored. Additionally, we define $F = \{f_1,f_2,...,f_N\}$ as the set of protected features that we aim to analyze and each $f_n$ has a set of navigation features $V$. To uncover social-group related bias, the next step is to determine the relationship degree $D_{k,n}$ between each protected feature $f_n$ and each generated cluster $A_k$. 

\begin{figure*}
\centering
\footnotesize
\includegraphics[width=0.8\textwidth]{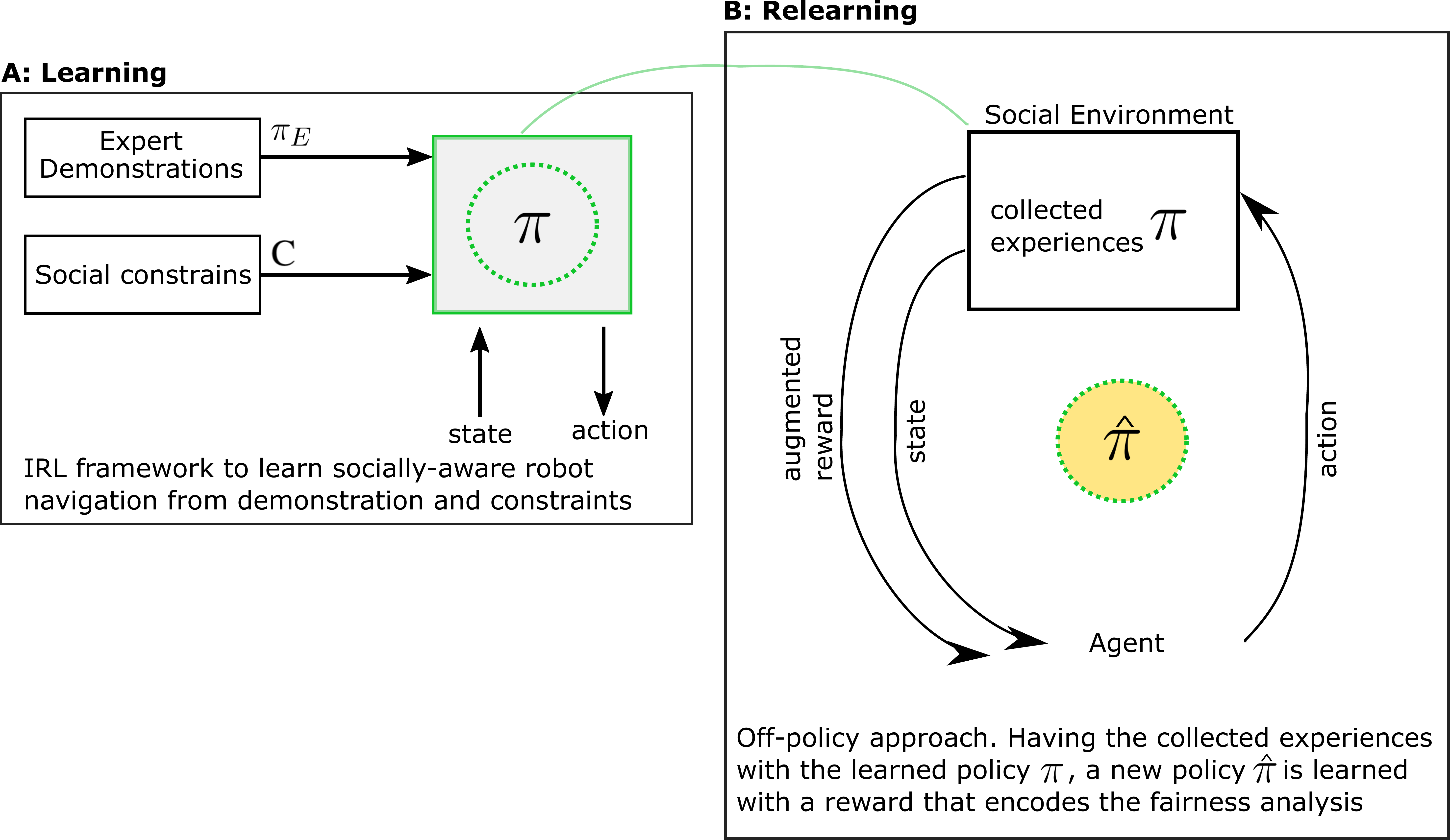}
\caption{Illustration of the Learning-Relearning framework for diminishing bias in social robot navigation. During the learning phase, a policy $\pi$ is learned for socially-aware robot navigation. During the relearning phase, the robot uses the policy $\pi$ to navigate in the social environment and collects the navigation experiences. An augmented reward that encodes detected biased behavior is used to relearn a new policy $\hat{\pi}$ so that the long term rewards reflect the decreasing unjustified bias related to social-groups.}
\label{fig:scheme2}
\end{figure*}

After identifying that the robot actions in the navigation experience set are clustered and correlated to sensitive attributes, the next step is to trigger alarms or corrective actions when protected feature $f_n$ strongly related to each generated cluster $A_k$, defined as $D_{k,n} > u_n$ where $u_n$ threshold that can be selected for each protected feature. A system of reward or punishment can be implemented in an off-policy reinforcement learning algorithm that optimizes an augmented reward that encodes the detection of unfair behavior as shown in Figure \ref{fig:scheme2}. The augmented reward $ {rR} _t $ is penalized when a biased behavior is detected, so it does not only comprise the behavior for socially-aware navigation but it is also discounted when we detect bias as $ D_{k, n }> u_n $. Therefore, the robot learns the policy ${\pi_R} $ so that the long term rewards reflect the decreasing unjustified bias related to social-groups. As a result, it is possible to relearn the navigation model in our framework depending on the information gathered from the social environment.

From a more realistic perspective, demographic information is rarely known. Clustering also allows the reduction of this dependency between predictions and demographic information, when an unsupervised approach is employed. Therefore, when the dataset containing memory experiences of the robot navigating conforms to clusters beyond a given threshold, it can trigger an alarm for further analysis. Other methodologies that can be used to undercover bias in deep learning models are based on visualization of embeddings. Using visualization techniques, we can show how the model groups the data, which is useful to expose the reasons behind the prediction of the model. To do so, different tools can be used, such as T-distributed stochastic neighbor embedding (t-SNE) and Uniform Manifold Approximation and Projection (UMAP), to project the embeddings to reduce the dimensionality of the data. In this work, we focus on the relearning component based on clustering to present a feasible solution to account for fairness while learning socially compliant robot navigation that can be extended to an unsupervised algorithm.

\section{Case Studies and Discussion}
\label{sec:case_study}

In this section, we present extensive discussions that relate the technical analysis of our proposed framework to complex real-world scenarios that we present as three case studies. Each of these case studies contains different levels of human-robot interaction under four specific protected characteristics: gender, disabilities, age, and race. With these scenarios, we analyze the feasibility of model adaptation and the utility of this mechanism to check for fairness as well as to correct the bias. The figures illustrated in this section were generated using~\cite{icograms}. 


\subsection{Autonomous Floor Cleaning Robots}
\label{sec:cleaning}

\begin{figure*}[h]
\centering
\footnotesize
\includegraphics[width=0.7\textwidth]{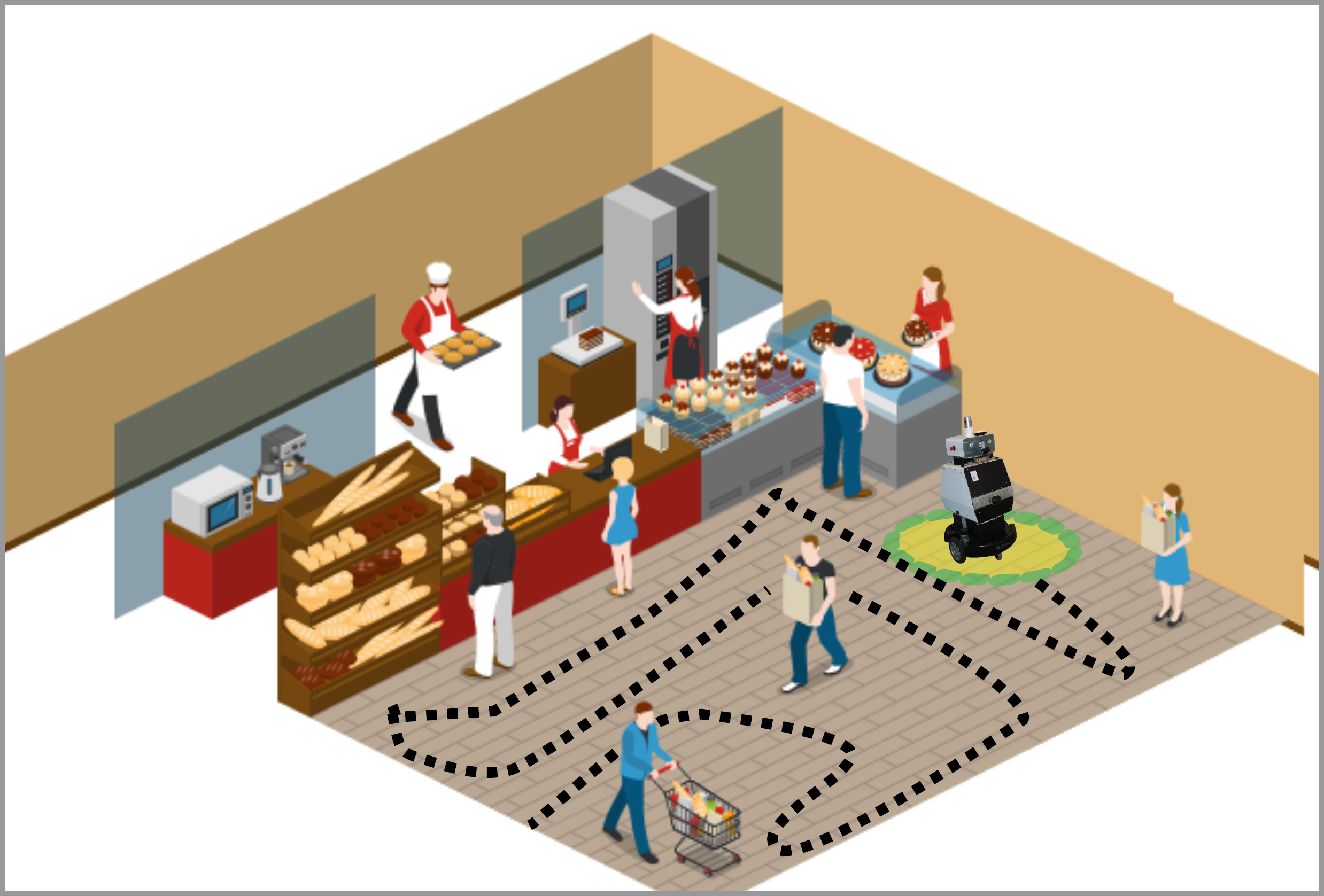}
\caption{Illustration of the autonomous floor cleaning robot scenario. The robot navigates taking the social conventions into account while performing the main task of cleaning the entire area.}
\label{fig:cs1}
\end{figure*}

One of the most societally accepted robots has been the autonomous floor-cleaning machines~\citep{fink2013living,forlizzi2006service,forlizzi2007robotic} and during the last decade, they have been the most sold robots in the world~\citep{clearningMarket}. These robots have the task of cleaning floors using vacuum systems without any human supervision, and recently, they can also mop floors using steam systems. These robots are currently used in households, and their navigation models vary in complexity depending on a wide range of prices. However, these robots are so far not equipped with socially aware navigation models. They do not avoid people or dynamic objects, rather they only change their cleaning route after they collide with an object. This can be attributed to the fact that in household environments, people are typically more tolerant given that they are aware of the task, features, and capacity of the robot. 

It can be expected that the use of cleaning robots in the future will spread to different public areas. In this case study, we analyze from both technological and social points of view the functioning, requirements, and implications of the navigation of a cleaning robot that operates in a shopping mall. We illustrate this scenario in Figure~\ref{fig:cs1}. Consider that the shopping mall consists of multiple and extensive floors, and it is open to the public continually every day of the week. The groups of people visiting the place range from families and groups of friends to individual persons. Additionally, the reasons for the visit can differ, including people making quick shops, taking a walk, eating, etc. Therefore, we also expect varying types of behavior of the visitors, such as walking at different speeds, talking in groups, and sitting down in different spaces.

The task of the robot in this case is to clean the entire environment effectively. In the following, we examine the effect that a cleaning robot equipped with social context can have. This robot has the ability to plan paths taking into account social conventions in public spaces such as avoiding interfering with the paths of people, avoiding interrupting the interaction between people, prioritizing safety, avoiding surprising people with movements outside the visual range (or any other movement that might make people uncomfortable), navigating with a safe distance and with a prudent speed, avoiding collisions and predicting the intentions of people. With socially-aware navigation models, robots can fulfill the main task and act socially with predictable actions. The goal of including social context into the navigation model is to ensure that robots are not perceived as dangerous, bothersome, irritating, inconvenient, or obtrusive. The sociability of the cleaning robots can be defined as low or indirect, i.e., humans do not communicate with the robot. However, the interaction is generated by the navigation model in a socially acceptable manner. Social navigation models allow the robot to achieve the main goal without disturbing people sharing the same space. Consequently, the robot can operate in public spaces during the entire opening hours. 

Specifically, if we employ the model~\citep{patompak2019learning} presented in Section~\ref{subsec:Tech_Fair} as the \textit{learning} component in our framework, the personalized size, and shape of the personal zone can in fact improve the social intelligence of the robot. By avoiding crossing the comfort zone of people, these robots can learn to plan paths without disturbing the visitors of the shopping mall while performing the cleaning task. However, the model~\citep{patompak2019learning} that takes the gender of a person into account can induce bias in the decisions. Even though women might prefer a larger comfort area during interaction among humans, it does not necessarily imply that they would prefer the same during human-robot interaction. In principle, a robot should never harm or be unfair to people based on their gender. In this work, we consider that the robot is depicted as a gender-neutral machine. Conforming a robot to a specific gender depending on the application could again lead to historical bias, this is an area that requires further research, which is out of the scope of this paper.
Moreover, according to the \textit{bias evaluation} consideration for fairness described in Section~\ref{sec:fairnessmeasures}, maintaining different relative distances to people based on their gender is an unacceptable bias. Furthermore, distinguishing the comfort area by gender is not of high relevance to improve the acceptance or beneficial to improve the operation of robots around humans. Instead, there are other essential factors that can be used to improve comfort and confidence, such as safe navigation policies. Given that the bias presented in this case is explicit, it is easier to identify the bias inducing factor influencing the model in the  \textit{relearning} component of our framework, for example, by correlating the obtained behavior to the input constraints. After detecting the bias inducing factor, it can be excluded to re-train the model without the gender constraint.

On the other hand, while learning from demonstrations, data-driven models can also reflect negative bias. For instance, if robots learn from data that is not diverse where people with movement impairments are not present, then the robot might not react in a socially acceptable manner when they encounter such people. This can further lead to an incorrect prediction of paths of people who walk slower and can make the robot be perceived as obtrusive. Data induced bias represents an implicit bias in the model that is more challenging to detect and correct for. Since the model disproportionately affects a specific group of people, by using our \textit{relearning} component, the recurrent errors in the path prediction can be detected as a cluster that can also be related to the set of protected characteristics (e.g., people with mobility impairment). Consequently, by using a punishment system, the reward value is influenced after the detection of unwanted behavior to adjust the learning policy, allowing model adaptation towards a more fair behavior. This will support the \textit{Value Alignment} consideration presented in Section~\ref{sec:fairnessmeasures} in which accepted socially-aware robot navigation also considers inclusion.

\subsection{Guidance Robots in a Shopping Mall}

\begin{figure*}[h]
\centering
\footnotesize
\includegraphics[width=0.7\textwidth]{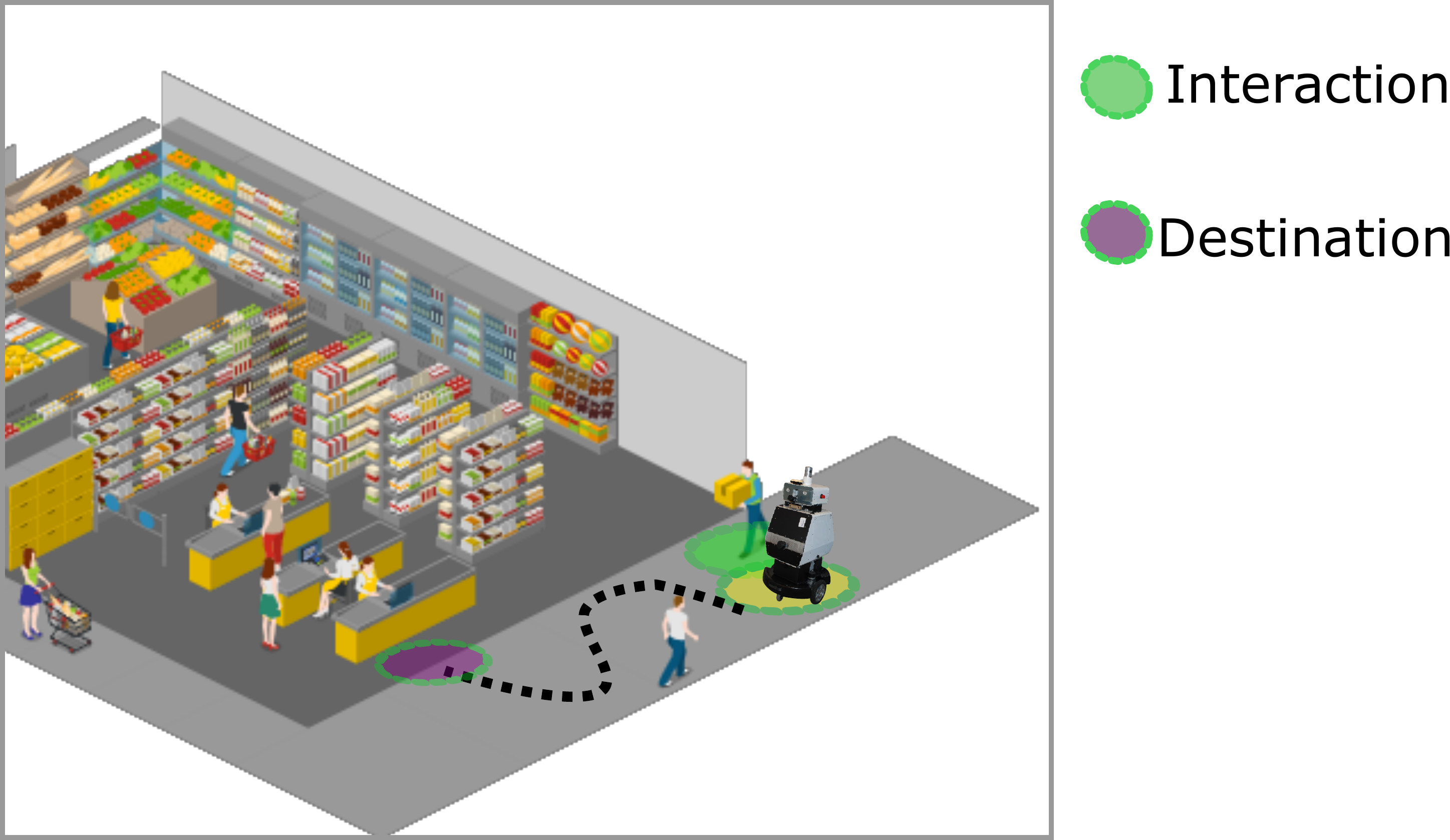}
\caption{Illustration of the guidance robot in a shopping mall scenario. The robot guides the user (green circle) to reach the destination (purple circle). Additionally, the robot is aware of the people in the surroundings during navigation while maintaining a desired relative position with respect to the user.}
\label{fig:cs2}
\end{figure*}

Mobile service robots have extensive use in innovative applications such as for guidance in public spaces where they navigate alongside people and assist them to reach their desired destination. Based on the environment described in Section~\ref{sec:cleaning}, in this case study we analyze the effects of a guidance robot that operates in a shopping mall. Unlike the last scenario, the robot not only navigates under social conventions but also guides a person in a social manner. The task of the robot is to provide the requested information about locations in the shopping mall and accompany people to reach their desired location. This scenario is illustrated in Figure~\ref{fig:cs2}. Apart from guiding to reach a certain destination, the robot should also navigate considering social conventions that are required to provide comfort to all the surrounding people during navigation. Furthermore, the robot should coordinate with the user while navigating by maintaining a desired relative position with respect to the user. This scenario has similar characteristics to the mall in the previous case study where diverse people with different genders, ethnicity, disabilities, age, skin tones, and cultural origins, will be present. In this example, fairness considerations such as shared benefit, deterrence and value alignment described in the Section~\ref{sec:fairnessmeasures} should be considered. Additionally, in the shopping mall scenario, the guidance robot will interact naturally with the user in a socially compliant manner while providing information and route guidance.

The human-robot interaction in this case is direct given that people approach the robot with a specific intention, and they expect a response from the robot that corresponds to the request. The resulting navigation strategy that these robots have next to people and their capacity to react according to the situation is crucial for their acceptance. Some of the important constraints in the navigation behavior of guidance robots are adapting the speed of the robot to the user, and maintaining a relative position and distance. If the robot navigates with a velocity that does not correspond to the user, then the robot risks being too slow or too fast which can cause uncoordinated behavior with the user and can further lead to accidents. On the other hand, relative distance and position are related to how people follow the robot and how the robot guides the user. Ideally, the robot should estimate the position and intention of the user during the execution of the guidance and also be able to interrupt the task if the person does not require any more help. Therefore, robots should adapt their navigation based on speed, intentions, motivations, orientation as well as handle unexpected situations such as people crossing their path, changes in the speed of the person being guided, unexpected appearance of objects, among others. 

Consumers value the unbiased, fast, and error-free behavior that a robot can provide. Therefore, the robot should adapt its behavior according to the current social context. In contrast to the interaction between people and cleaning robots, guidance robots provide personalized interaction, so the degree of sociability of this robot is greater. For example, if a disabled person goes to a shopping mall, the robot should recognize that this person will have different navigation behaviors than others so it should adapt its strategy accordingly. This adaptation will, in turn, make the person more comfortable using the assistance provided by the robot. In this example, aspects such as the capability to recognize mobility impairments in a person and navigate accordingly are essential to ensure safe and comfortable guidance. Consider that a person with limited mobility requires guidance from the robot. If the robot is not equipped to react accordingly to mobility difficulties, the interaction can cause distress, physical overexertion, and even accidents. This will eventually make the person discontinue using the robot in the future. In order to avoid such events, the navigation model in the robot should incorporate social adaptability skills that enable it to detect particular situations that cause discomfort or unintended outcomes for specific individuals.

Assume that a guidance robot is equipped with the navigation model described in Section~\ref{subsec:Tech_Fair} and as a consequence, it will assist women keeping larger distances with them. This may cause the robot to lose the interaction with them in certain situations and adversely affect the way that women perceive the robot. Similarly, it can reduce the efficiency with this population group representing the systematic disadvantage we aim to avoid towards diminishing bias. The model described in Section~\ref{subsec:Tech_Fair} is used to present an example of learning socially-aware robot navigation in which unfair outcomes are associated with a protected characteristic. Other socially-aware navigation models that learn solely from human imitation can cause different types of model-induced biases. In these cases, the navigation model is optimized to yield sociable actions considering different factors such as the velocity, orientation, priority of interaction, and route selection. The guidance robot will encounter situations where multiple people request for help simultaneously or even situations where people will try to interact with the robot when it is already guiding another person. Deciding which person has the priority is part of the social intelligence. Assume that in the \textit{learning} component of our framework, the navigation model of the robot is trained from demonstrations, and as a result, the robot learns the preferred interaction behavior based on those demonstrated interactions. This can lead to unfair outcomes due to human bias that may be existing in the demonstrations, policies reflecting personal bias, unequal society roles, or under-representation of minorities. Specifically, if the learning from demonstration is performed in a shopping mall only from one city, there will be insufficient diversity. Similarly, if the robot is deployed in a different place, or when people belonging to minorities try to use the robot, the robot will maintain its social behavior, but it will likely make biased decisions, especially against people who historically have been discriminated against, as we observed in other cases~\citep{brandao2019age,buolamwini2018gender,wilson2019predictive,prabhu2020large}. As part of the relearning component, our framework allows us to generate clusters related to preferred interaction actions and determine if the generated clusters are strongly related to protected characteristics. Specifically, in case the preferred interaction of the robot is biased favoring or disadvantaging specific visitors of the shopping mall the learning policy is adjusted by a reward value that is penalized when biased behavior is detected. As a consequence, the robot's actions, such as deciding which person has the priority to interact with will follow the fairness requirements.

Since diverse people typically visit shopping malls, the robot should be able to accurately recognize them regardless of factors such as skin tones. Previous studies~\citep{wilson2019predictive} have shown that recognition systems based on RGB perception present higher error rates for dark skin tones. If similar systems with faulty sensors or algorithms are used to learn social navigation models, the robot will be unable to recognize certain people and adhere to the fairness considerations described in Section~\ref{sec:fairnessmeasures}. As a consequence, the robot can perpetuate discrimination against groups of people that have historically been segregated, as observed in other learning applications such as the automated risk assessment used by U.S judges and the biased vision-based object detectors employed in autonomous cars~\citep{benthall2019racial,wilson2019predictive}. Furthermore, discrimination laws prohibit the unfair treatment of people based on race. In this case, fairness priority is also important for the legal framework.

 \subsection{Caregiving Robots in Hospitals}

\begin{figure*}[h]
\centering
\footnotesize
\includegraphics[width=0.7\textwidth]{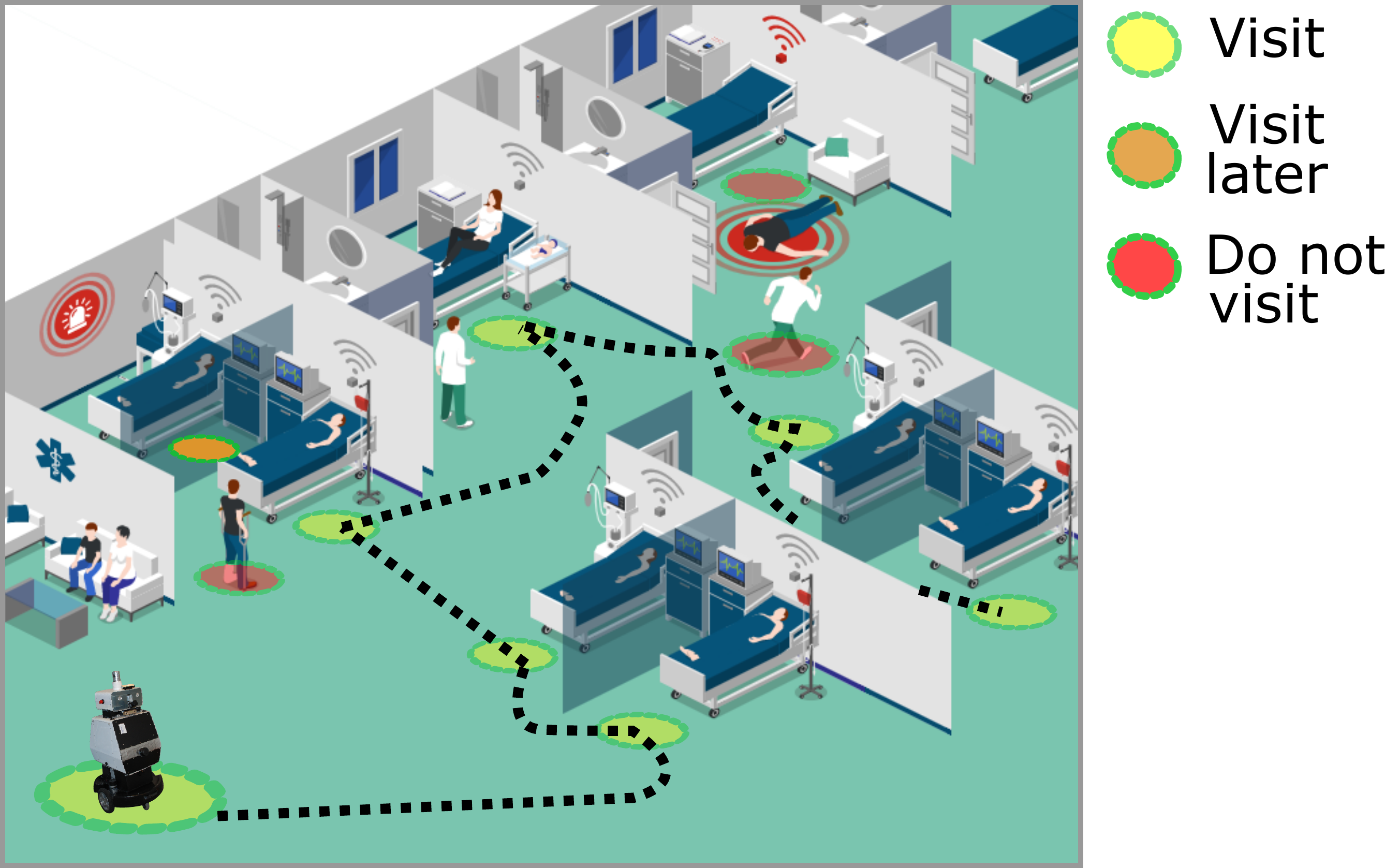}
\caption{Illustration of the caregiving robot in a hospital scenario. The main task of the robot is to distribute medicines to patients who are admitted in the hospital. The robot takes emergency situations that could happen into account and people requiring special assistance, while navigating.}
\label{fig:cs3}
\end{figure*}

There is significant interest in developing service robots for hospitals due to their ability to provide care for people. The use of robots in hospitals can be especially advantageous in cases where there are patients with contagious diseases, such as in a pandemic situation. In this case study, we analyze the navigation strategy of caregiving robots that operate in hospitals. The main task of robots in this case study is to distribute medicines to patients who are admitted in a hospital. Figure~\ref{fig:cs3} illustrates this scenario. The human-robot interaction in hospitals requires special caution as the robot will operate around patients who require special assistance. One such example is people with motion impairments who use wheelchairs, crutches, or walking frames. Furthermore, the robot will encounter rapidly changing situations, for example, during an emergency where doctors and care staff rush through the hallways. To provide appropriate response, robots should be equipped with algorithms to understand situations and context that enable them to accordingly adapt their behavior. Apart from patients, robots will also interact with other people in the hospital, such as health professionals, secretaries, family members, and visitors. Similar to the shopping mall case study, caregiving robots will be interacting directly with the people. However, the navigation and interaction present additional complexity, given that they do not assist people individually. Here, the robots aim to assist multiple people who have different medical treatments and deliver medicine to them while maintaining a socially accepted behavior. In this case, not only social conventions and sociability described in the previous case studies are required, but also priority decision making, optimal recognition, faster reaction, and adaptability. As a consequence, the navigation models in caregiving robots should have higher requirements of accuracy and adaptability. These robots can particularly encounter unexpected events such as emergency situations where people will be walking in different directions, speeds, and unpredictable movements. In such situations, there is a higher risk of accidents due to the vulnerability of people and the context in the hospital. Furthermore, the consequences of eventual accidents can be critical for the health of individuals. Caregiving robots should be able to perceive, recognize, and react according to the special requirements of the hospital.

Assume that the robots are going to be used in emergency rooms. Their task there is to deliver a series of necessary supplies to the people who are attending to the emergencies. Therefore, the robots have to interact with several people simultaneously. Based on the proxemics model described in Section~\ref{subsec:Tech_Fair}, the robot will be perceived as atypical in approaching people in different ways, assisting some people differently than others during urgent situations. Furthermore, taking into account that there are people playing specific roles, namely to care for sick people urgently, their comfort area of interaction is different from that of normal situations. People typically tend to walk fast, to have little personal space, and to quickly perceive what is happening around them. In this scenario, robots that navigate while maintaining different distances to people based on gender have lesser foreseeable utility. Alternatively, other characteristics can be considered that are related to the distribution of medicines depending on the needs of the patients and priorities, such as minimizing delivery time.

The priority of the path planning algorithms in such robots is to deliver medicines to all patients. Assume that in the \textit{learning component} the caregiving robot learns from historical data about the characteristics of the patients. This model may learn that the pain threshold differs between men and women. Consequently, the navigation plan will be biased with negative effects towards men, based on information related to their higher tolerance to pain. Similarly, the robot could learn that women have more tolerance to wait longer for medical treatments and spend more overall time than men in the emergency rooms~\citep{nottingham2018effect}. In both situations, the behavior of the robot will be biased given that it systematically benefits a specific group of people. In this example, fairness considerations such as value alignment and non-maleficence described in the Section~\ref{sec:fairnessmeasures} can improve the decisions made by the robot. One approach to dealing with difficult cases of priority is to reflect political and commercial neutrality in robot navigation. This signifies that the navigation model in caregiving robots should not favor any particular group of people. Although advocating for the neutrality of assistive robots is a potential solution to bias problems, in this case, the concept is substantially complex and requires further research.

Particularly, adapting the model with our \textit{relearning} component to correct for the presented bias will lead the robot to base decisions on other factors. Using the \textit{relearning} component of our framework, we can identify clusters that demonstrate a systematic disadvantage if the time to deliver medicines is higher for men and if women wait for a longer period of time in emergency rooms. Subsequently, to penalize the unfair behavior, we lower the reward value that adjusts the learning policy. As a result, the navigation model is adapted towards more fair behavior. If the model does not rely on the potentially negative bias inducing factors, it can learn better representations that reflect relevant characteristics such as urgency and needs. While using our relearning technique, this type of bias in navigation will be detected when certain people receive attention more effectively than others. Consequently, if there is no valid reasoning behind such bias, the navigation model should be updated accordingly.



\section{Conclusions}

As more and more robots navigate in human spaces, they also require more complex navigation models to accomplish their goals while complying with the high safety and comfort requirements. Towards this direction, different methods incorporate social context into learning models to enable robots to navigate following social conventions. Typically, these methodologies utilize data or experiences from the real world, simulations, or control experiments and social constraints. In this work, we discussed the societal and ethical implications of learned socially-aware robot navigation techniques. We demonstrated that the advances accomplished in social robot navigation are essential for the development of robots that provide well for society. More importantly, we showed how these models that account for socially-aware robot navigation do not guarantee fairness in different real-world scenarios. Research in the direction of fairness in robot learning is of special importance, given that these machines interact with people closely.

To the best of our knowledge, this is the first work that studies the societal implications of bias in learned socially-aware robot navigation models. Our proposed framework that consists of the learning and relearning stages has the ability to effectively diminish bias in social robot navigation models. Additionally, we presented fairness considerations and specific techniques that can be used to implement our framework. We detailed several scenarios that show that the adaptability of the model in terms of fairness enables it to correct for bias. The scenarios demonstrate the potential unwanted outcomes of social navigation models that are described with variables and social conventions which make them easily interpretable. Our framework is especially useful for more complex learning models or models that are trained with imitation or reinforcement learning, given that these models contain more abstract representations of the data and situations. We hope this work contributes towards raising awareness of the importance of fairness in robot learning.



\section*{Acknowledgement}

This work was partly funded by the BrainLinks-BrainTools center of the University of Freiburg, a scholarship from the Graduate School of Robotics of the University Freiburg (according to the Graduate Funding Law of the Ministry of Science, Research and Arts of the State of Baden-Württemberg), and a grant from the Eva~Mayr-Stihl Stiftung.


\bibliographystyle{elsarticle-num-names}
\bibliography{hurtado21frai}

\begin{thebibliography}{114}
\expandafter\ifx\csname natexlab\endcsname\relax\def\natexlab#1{#1}\fi
\providecommand{\url}[1]{\texttt{#1}}
\providecommand{\href}[2]{#2}
\providecommand{\path}[1]{#1}
\providecommand{\DOIprefix}{doi:}
\providecommand{\ArXivprefix}{arXiv:}
\providecommand{\URLprefix}{URL: }
\providecommand{\Pubmedprefix}{pmid:}
\providecommand{\doi}[1]{\href{http://dx.doi.org/#1}{\path{#1}}}
\providecommand{\Pubmed}[1]{\href{pmid:#1}{\path{#1}}}
\providecommand{\bibinfo}[2]{#2}
\ifx\xfnm\relax \def\xfnm[#1]{\unskip,\space#1}\fi
\bibitem[{Tewari et~al.(2002)Tewari, Peabody, Sarle, Balakrishnan, Hemal,
  Shrivastava, and Menon}]{tewari2002technique}
\bibinfo{author}{A.~Tewari}, \bibinfo{author}{J.~Peabody},
  \bibinfo{author}{R.~Sarle}, \bibinfo{author}{G.~Balakrishnan},
  \bibinfo{author}{A.~Hemal}, \bibinfo{author}{A.~Shrivastava},
  \bibinfo{author}{M.~Menon},
\newblock \bibinfo{title}{Technique of da vinci robot-assisted anatomic radical
  prostatectomy},
\newblock \bibinfo{journal}{Urology} \bibinfo{volume}{60}
  (\bibinfo{year}{2002}) \bibinfo{pages}{569--572}.
\bibitem[{Thrun(1995)}]{thrun1995approach}
\bibinfo{author}{S.~Thrun},
\newblock \bibinfo{title}{An approach to learning mobile robot navigation},
\newblock \bibinfo{journal}{Robotics and Autonomous systems}
  \bibinfo{volume}{15} (\bibinfo{year}{1995}) \bibinfo{pages}{301--319}.
\bibitem[{Valada et~al.(2012)Valada, Tomaszewski, Kannan, Velagapudi, Kantor,
  and Scerri}]{valada2012intelligent}
\bibinfo{author}{A.~Valada}, \bibinfo{author}{C.~Tomaszewski},
  \bibinfo{author}{B.~Kannan}, \bibinfo{author}{P.~Velagapudi},
  \bibinfo{author}{G.~Kantor}, \bibinfo{author}{P.~Scerri},
\newblock \bibinfo{title}{An intelligent approach to hysteresis compensation
  while sampling using a fleet of autonomous watercraft},
\newblock in: \bibinfo{booktitle}{International Conference on Intelligent
  Robotics and Applications}, \bibinfo{organization}{Springer},
  \bibinfo{year}{2012}, pp. \bibinfo{pages}{472--485}.
\bibitem[{Toupet et~al.(2020)Toupet, Biesiadecki, Rankin, Steffy,
  Meirion-Griffith, Levine, Schadegg, and Maimone}]{toupet2020terrain}
\bibinfo{author}{O.~Toupet}, \bibinfo{author}{J.~Biesiadecki},
  \bibinfo{author}{A.~Rankin}, \bibinfo{author}{A.~Steffy},
  \bibinfo{author}{G.~Meirion-Griffith}, \bibinfo{author}{D.~Levine},
  \bibinfo{author}{M.~Schadegg}, \bibinfo{author}{M.~Maimone},
\newblock \bibinfo{title}{Terrain-adaptive wheel speed control on the curiosity
  mars rover: Algorithm and flight results},
\newblock \bibinfo{journal}{Journal of Field Robotics} \bibinfo{volume}{37}
  (\bibinfo{year}{2020}) \bibinfo{pages}{699--728}.
\bibitem[{Mittal et~al.(2019)Mittal, Mohan, Burgard, and
  Valada}]{mittal2019vision}
\bibinfo{author}{M.~Mittal}, \bibinfo{author}{R.~Mohan},
  \bibinfo{author}{W.~Burgard}, \bibinfo{author}{A.~Valada},
\newblock \bibinfo{title}{Vision-based autonomous uav navigation and landing
  for urban search and rescue},
\newblock \bibinfo{journal}{arXiv preprint arXiv:1906.01304}
  (\bibinfo{year}{2019}).
\bibitem[{Gaydashenko et~al.(2018)Gaydashenko, Kudenko, and
  Shpilman}]{gaydashenko2018comparative}
\bibinfo{author}{A.~Gaydashenko}, \bibinfo{author}{D.~Kudenko},
  \bibinfo{author}{A.~Shpilman},
\newblock \bibinfo{title}{A comparative evaluation of machine learning methods
  for robot navigation through human crowds},
\newblock in: \bibinfo{booktitle}{2018 17th IEEE International Conference on
  Machine Learning and Applications (ICMLA)}, \bibinfo{organization}{IEEE},
  \bibinfo{year}{2018}, pp. \bibinfo{pages}{553--557}.
\bibitem[{Boniardi et~al.(2016)Boniardi, Valada, Burgard, and
  Tipaldi}]{boniardi2016autonomous}
\bibinfo{author}{F.~Boniardi}, \bibinfo{author}{A.~Valada},
  \bibinfo{author}{W.~Burgard}, \bibinfo{author}{G.~D. Tipaldi},
\newblock \bibinfo{title}{Autonomous indoor robot navigation using sketched
  maps and routes},
\newblock \bibinfo{journal}{Workshop on Model Learning for Human-Robot
  Communication at Robotics: Science and Systems (RSS)}
  (\bibinfo{year}{2016}).
\bibitem[{Jamshidi et~al.(2019)Jamshidi, C{\'a}mara, Schmerl, K{\"a}estner, and
  Garlan}]{jamshidi2019machine}
\bibinfo{author}{P.~Jamshidi}, \bibinfo{author}{J.~C{\'a}mara},
  \bibinfo{author}{B.~Schmerl}, \bibinfo{author}{C.~K{\"a}estner},
  \bibinfo{author}{D.~Garlan},
\newblock \bibinfo{title}{Machine learning meets quantitative planning:
  Enabling self-adaptation in autonomous robots},
\newblock in: \bibinfo{booktitle}{2019 IEEE/ACM 14th International Symposium on
  Software Engineering for Adaptive and Self-Managing Systems (SEAMS)},
  \bibinfo{organization}{IEEE}, \bibinfo{year}{2019}, pp.
  \bibinfo{pages}{39--50}.
\bibitem[{Hurtado et~al.(2020)Hurtado, Mohan, and Valada}]{hurtado2020mopt}
\bibinfo{author}{J.~V. Hurtado}, \bibinfo{author}{R.~Mohan},
  \bibinfo{author}{A.~Valada},
\newblock \bibinfo{title}{Mopt: Multi-object panoptic tracking},
\newblock \bibinfo{journal}{arXiv preprint arXiv:2004.08189}
  (\bibinfo{year}{2020}).
\bibitem[{Nolfi and Floreano(2002)}]{nolfi2002synthesis}
\bibinfo{author}{S.~Nolfi}, \bibinfo{author}{D.~Floreano},
\newblock \bibinfo{title}{Synthesis of autonomous robots through evolution},
\newblock \bibinfo{journal}{Trends in cognitive sciences} \bibinfo{volume}{6}
  (\bibinfo{year}{2002}) \bibinfo{pages}{31--37}.
\bibitem[{Poudel(2013)}]{poudel2013coordinating}
\bibinfo{author}{D.~B. Poudel},
\newblock \bibinfo{title}{Coordinating hundreds of cooperative, autonomous
  robots in a warehouse},
\newblock \bibinfo{journal}{Jan} \bibinfo{volume}{27} (\bibinfo{year}{2013})
  \bibinfo{pages}{1--13}.
\bibitem[{Bogue(2016)}]{bogue2016search}
\bibinfo{author}{R.~Bogue},
\newblock \bibinfo{title}{Search and rescue and disaster relief robots: has
  their time finally come?},
\newblock \bibinfo{journal}{Industrial Robot: An International Journal}
  (\bibinfo{year}{2016}).
\bibitem[{Hasan et~al.(2014)Hasan, Reza et~al.}]{hasan2014path}
\bibinfo{author}{K.~M. Hasan}, \bibinfo{author}{K.~J. Reza}, et~al.,
\newblock \bibinfo{title}{Path planning algorithm development for autonomous
  vacuum cleaner robots},
\newblock in: \bibinfo{booktitle}{2014 International Conference on Informatics,
  Electronics \& Vision (ICIEV)}, \bibinfo{organization}{IEEE},
  \bibinfo{year}{2014}, pp. \bibinfo{pages}{1--6}.
\bibitem[{Khambhaita and Alami(2020)}]{khambhaita2020viewing}
\bibinfo{author}{H.~Khambhaita}, \bibinfo{author}{R.~Alami},
\newblock \bibinfo{title}{Viewing robot navigation in human environment as a
  cooperative activity},
\newblock in: \bibinfo{booktitle}{Robotics Research},
  \bibinfo{publisher}{Springer}, \bibinfo{year}{2020}, pp.
  \bibinfo{pages}{285--300}.
\bibitem[{Wittrock(2010)}]{wittrock2010learning}
\bibinfo{author}{M.~C. Wittrock},
\newblock \bibinfo{title}{Learning as a generative process},
\newblock \bibinfo{journal}{Educational Psychologist} \bibinfo{volume}{45}
  (\bibinfo{year}{2010}) \bibinfo{pages}{40--45}.
\bibitem[{Bicchi and Tamburrini(2015)}]{bicchi2015social}
\bibinfo{author}{A.~Bicchi}, \bibinfo{author}{G.~Tamburrini},
\newblock \bibinfo{title}{Social robotics and societies of robots},
\newblock \bibinfo{journal}{The Information Society} \bibinfo{volume}{31}
  (\bibinfo{year}{2015}) \bibinfo{pages}{237--243}.
\bibitem[{Silver et~al.(2010)Silver, Bagnell, and Stentz}]{silver2010learning}
\bibinfo{author}{D.~Silver}, \bibinfo{author}{J.~A. Bagnell},
  \bibinfo{author}{A.~Stentz},
\newblock \bibinfo{title}{Learning from demonstration for autonomous navigation
  in complex unstructured terrain},
\newblock \bibinfo{journal}{The International Journal of Robotics Research}
  \bibinfo{volume}{29} (\bibinfo{year}{2010}) \bibinfo{pages}{1565--1592}.
\bibitem[{Kretzschmar et~al.(2016)Kretzschmar, Spies, Sprunk, and
  Burgard}]{kretzschmar2016socially}
\bibinfo{author}{H.~Kretzschmar}, \bibinfo{author}{M.~Spies},
  \bibinfo{author}{C.~Sprunk}, \bibinfo{author}{W.~Burgard},
\newblock \bibinfo{title}{Socially compliant mobile robot navigation via
  inverse reinforcement learning},
\newblock \bibinfo{journal}{The International Journal of Robotics Research}
  \bibinfo{volume}{35} (\bibinfo{year}{2016}) \bibinfo{pages}{1289--1307}.
\bibitem[{Luber et~al.(2012)Luber, Spinello, Silva, and
  Arras}]{luber2012socially}
\bibinfo{author}{M.~Luber}, \bibinfo{author}{L.~Spinello},
  \bibinfo{author}{J.~Silva}, \bibinfo{author}{K.~O. Arras},
\newblock \bibinfo{title}{Socially-aware robot navigation: A learning
  approach},
\newblock in: \bibinfo{booktitle}{2012 IEEE/RSJ International Conference on
  Intelligent Robots and Systems}, \bibinfo{organization}{IEEE},
  \bibinfo{year}{2012}, pp. \bibinfo{pages}{902--907}.
\bibitem[{Ferrer et~al.(2013)Ferrer, Garrell, and Sanfeliu}]{ferrer2013social}
\bibinfo{author}{G.~Ferrer}, \bibinfo{author}{A.~Garrell},
  \bibinfo{author}{A.~Sanfeliu},
\newblock \bibinfo{title}{Social-aware robot navigation in urban environments},
\newblock in: \bibinfo{booktitle}{2013 European Conference on Mobile Robots},
  \bibinfo{organization}{IEEE}, \bibinfo{year}{2013}, pp.
  \bibinfo{pages}{331--336}.
\bibitem[{Kruse et~al.(2013)Kruse, Pandey, Alami, and Kirsch}]{kruse2013human}
\bibinfo{author}{T.~Kruse}, \bibinfo{author}{A.~K. Pandey},
  \bibinfo{author}{R.~Alami}, \bibinfo{author}{A.~Kirsch},
\newblock \bibinfo{title}{Human-aware robot navigation: A survey},
\newblock \bibinfo{journal}{Robotics and Autonomous Systems}
  \bibinfo{volume}{61} (\bibinfo{year}{2013}) \bibinfo{pages}{1726--1743}.
\bibitem[{Buolamwini and Gebru(2018)}]{buolamwini2018gender}
\bibinfo{author}{J.~Buolamwini}, \bibinfo{author}{T.~Gebru},
\newblock \bibinfo{title}{Gender shades: Intersectional accuracy disparities in
  commercial gender classification},
\newblock in: \bibinfo{booktitle}{Conference on fairness, accountability and
  transparency}, \bibinfo{year}{2018}, pp. \bibinfo{pages}{77--91}.
\bibitem[{Garcia(2016)}]{garcia2016racist}
\bibinfo{author}{M.~Garcia},
\newblock \bibinfo{title}{Racist in the machine: The disturbing implications of
  algorithmic bias},
\newblock \bibinfo{journal}{World Policy Journal} \bibinfo{volume}{33}
  (\bibinfo{year}{2016}) \bibinfo{pages}{111--117}.
\bibitem[{Wang et~al.(2020)Wang, Xu, Chen, Wang, Liu, and Qu}]{wang2020visual}
\bibinfo{author}{Q.~Wang}, \bibinfo{author}{Z.~Xu}, \bibinfo{author}{Z.~Chen},
  \bibinfo{author}{Y.~Wang}, \bibinfo{author}{S.~Liu}, \bibinfo{author}{H.~Qu},
\newblock \bibinfo{title}{Visual analysis of discrimination in machine
  learning},
\newblock \bibinfo{journal}{IEEE Transactions on Visualization and Computer
  Graphics}  (\bibinfo{year}{2020}).
\bibitem[{Lu et~al.(2020)Lu, Mardziel, Wu, Amancharla, and
  Datta}]{lu2020gender}
\bibinfo{author}{K.~Lu}, \bibinfo{author}{P.~Mardziel},
  \bibinfo{author}{F.~Wu}, \bibinfo{author}{P.~Amancharla},
  \bibinfo{author}{A.~Datta},
\newblock \bibinfo{title}{Gender bias in neural natural language processing},
\newblock in: \bibinfo{booktitle}{Logic, Language, and Security},
  \bibinfo{publisher}{Springer}, \bibinfo{year}{2020}, pp.
  \bibinfo{pages}{189--202}.
\bibitem[{Costa-juss{\`a}(2019)}]{costa2019analysis}
\bibinfo{author}{M.~R. Costa-juss{\`a}},
\newblock \bibinfo{title}{An analysis of gender bias studies in natural
  language processing},
\newblock \bibinfo{journal}{Nature Machine Intelligence}
  (\bibinfo{year}{2019}) \bibinfo{pages}{1--2}.
\bibitem[{Benthall and Haynes(2019)}]{benthall2019racial}
\bibinfo{author}{S.~Benthall}, \bibinfo{author}{B.~D. Haynes},
\newblock \bibinfo{title}{Racial categories in machine learning},
\newblock in: \bibinfo{booktitle}{Proceedings of the Conference on Fairness,
  Accountability, and Transparency}, \bibinfo{year}{2019}, pp.
  \bibinfo{pages}{289--298}.
\bibitem[{Wilson et~al.(2019)Wilson, Hoffman, and
  Morgenstern}]{wilson2019predictive}
\bibinfo{author}{B.~Wilson}, \bibinfo{author}{J.~Hoffman},
  \bibinfo{author}{J.~Morgenstern},
\newblock \bibinfo{title}{Predictive inequity in object detection},
\newblock \bibinfo{journal}{arXiv preprint arXiv:1902.11097}
  (\bibinfo{year}{2019}).
\bibitem[{Agarwal et~al.(2018)Agarwal, Beygelzimer, Dud{\'\i}k, Langford, and
  Wallach}]{agarwal2018reductions}
\bibinfo{author}{A.~Agarwal}, \bibinfo{author}{A.~Beygelzimer},
  \bibinfo{author}{M.~Dud{\'\i}k}, \bibinfo{author}{J.~Langford},
  \bibinfo{author}{H.~Wallach},
\newblock \bibinfo{title}{A reductions approach to fair classification},
\newblock \bibinfo{journal}{arXiv preprint arXiv:1803.02453}
  (\bibinfo{year}{2018}).
\bibitem[{Zafar et~al.(2017)Zafar, Valera, Rogriguez, and
  Gummadi}]{zafar2017fairness}
\bibinfo{author}{M.~B. Zafar}, \bibinfo{author}{I.~Valera},
  \bibinfo{author}{M.~G. Rogriguez}, \bibinfo{author}{K.~P. Gummadi},
\newblock \bibinfo{title}{Fairness constraints: Mechanisms for fair
  classification},
\newblock in: \bibinfo{booktitle}{Artificial Intelligence and Statistics},
  \bibinfo{organization}{PMLR}, \bibinfo{year}{2017}, pp.
  \bibinfo{pages}{962--970}.
\bibitem[{Dixon et~al.(2018)Dixon, Li, Sorensen, Thain, and
  Vasserman}]{dixon2018measuring}
\bibinfo{author}{L.~Dixon}, \bibinfo{author}{J.~Li},
  \bibinfo{author}{J.~Sorensen}, \bibinfo{author}{N.~Thain},
  \bibinfo{author}{L.~Vasserman},
\newblock \bibinfo{title}{Measuring and mitigating unintended bias in text
  classification},
\newblock in: \bibinfo{booktitle}{Proceedings of the 2018 AAAI/ACM Conference
  on AI, Ethics, and Society}, \bibinfo{year}{2018}, pp.
  \bibinfo{pages}{67--73}.
\bibitem[{Woodworth et~al.(2017)Woodworth, Gunasekar, Ohannessian, and
  Srebro}]{woodworth2017learning}
\bibinfo{author}{B.~Woodworth}, \bibinfo{author}{S.~Gunasekar},
  \bibinfo{author}{M.~I. Ohannessian}, \bibinfo{author}{N.~Srebro},
\newblock \bibinfo{title}{Learning non-discriminatory predictors},
\newblock \bibinfo{journal}{arXiv preprint arXiv:1702.06081}
  (\bibinfo{year}{2017}).
\bibitem[{McDonald et~al.(2008)McDonald, McCarthy, Soroczak, Nguyen, and
  Rashid}]{mcdonald2008proactive}
\bibinfo{author}{D.~W. McDonald}, \bibinfo{author}{J.~F. McCarthy},
  \bibinfo{author}{S.~Soroczak}, \bibinfo{author}{D.~H. Nguyen},
  \bibinfo{author}{A.~M. Rashid},
\newblock \bibinfo{title}{Proactive displays: Supporting awareness in fluid
  social environments},
\newblock \bibinfo{journal}{ACM Transactions on Computer-Human Interaction
  (TOCHI)} \bibinfo{volume}{14} (\bibinfo{year}{2008}) \bibinfo{pages}{1--31}.
\bibitem[{Hutchins(2006)}]{hutchins2006distributed}
\bibinfo{author}{E.~Hutchins},
\newblock \bibinfo{title}{The distributed cognition perspective on human
  interaction},
\newblock \bibinfo{journal}{Roots of human sociality: Culture, cognition and
  interaction} \bibinfo{volume}{1} (\bibinfo{year}{2006}) \bibinfo{pages}{375}.
\bibitem[{Goodwin(2000)}]{goodwin2000action}
\bibinfo{author}{C.~Goodwin},
\newblock \bibinfo{title}{Action and embodiment within situated human
  interaction},
\newblock \bibinfo{journal}{Journal of pragmatics} \bibinfo{volume}{32}
  (\bibinfo{year}{2000}) \bibinfo{pages}{1489--1522}.
\bibitem[{Jarvis(2006)}]{jarvis2006towards}
\bibinfo{author}{P.~Jarvis}, \bibinfo{title}{Towards a comprehensive theory of
  human learning}, volume~\bibinfo{volume}{1}, \bibinfo{publisher}{Psychology
  Press}, \bibinfo{year}{2006}.
\bibitem[{Johnson and Kuipers(2018)}]{johnson2018socially}
\bibinfo{author}{C.~Johnson}, \bibinfo{author}{B.~Kuipers},
\newblock \bibinfo{title}{Socially-aware navigation using topological maps and
  social norm learning},
\newblock in: \bibinfo{booktitle}{Proceedings of the 2018 AAAI/ACM Conference
  on AI, Ethics, and Society}, \bibinfo{year}{2018}, pp.
  \bibinfo{pages}{151--157}.
\bibitem[{Ulrich and Borenstein(2001)}]{ulrich2001guidecane}
\bibinfo{author}{I.~Ulrich}, \bibinfo{author}{J.~Borenstein},
\newblock \bibinfo{title}{The guidecane-applying mobile robot technologies to
  assist the visually impaired},
\newblock \bibinfo{journal}{IEEE Transactions on Systems, Man, and
  Cybernetics-Part A: Systems and Humans} \bibinfo{volume}{31}
  (\bibinfo{year}{2001}) \bibinfo{pages}{131--136}.
\bibitem[{Hagendorff(2020)}]{hagendorff2020ethical}
\bibinfo{author}{T.~Hagendorff},
\newblock \bibinfo{title}{Ethical behavior in humans and machines--evaluating
  training data quality for beneficial machine learning},
\newblock \bibinfo{journal}{arXiv preprint arXiv:2008.11463}
  (\bibinfo{year}{2020}).
\bibitem[{Piano(2020)}]{piano2020ethical}
\bibinfo{author}{S.~L. Piano},
\newblock \bibinfo{title}{Ethical principles in machine learning and artificial
  intelligence: cases from the field and possible ways forward},
\newblock \bibinfo{journal}{Humanities and Social Sciences Communications}
  \bibinfo{volume}{7} (\bibinfo{year}{2020}) \bibinfo{pages}{1--7}.
\bibitem[{Silberg and Manyika(2019)}]{silberg2019notes}
\bibinfo{author}{J.~Silberg}, \bibinfo{author}{J.~Manyika},
\newblock \bibinfo{title}{Notes from the ai frontier: Tackling bias in ai (and
  in humans)},
\newblock \bibinfo{journal}{Mckinsey Global Inst., Jun}
  (\bibinfo{year}{2019}).
\bibitem[{Cath(2018)}]{cath2018governing}
\bibinfo{author}{C.~Cath}, \bibinfo{title}{Governing artificial intelligence:
  ethical, legal and technical opportunities and challenges},
  \bibinfo{year}{2018}.
\bibitem[{Liu and Zawieska(2017)}]{liu2017responsible}
\bibinfo{author}{H.-Y. Liu}, \bibinfo{author}{K.~Zawieska},
\newblock \bibinfo{title}{From responsible robotics towards a human rights
  regime oriented to the challenges of robotics and artificial intelligence},
\newblock \bibinfo{journal}{Ethics and Information Technology}
  (\bibinfo{year}{2017}) \bibinfo{pages}{1--13}.
\bibitem[{Birhane and Cummins(2019)}]{birhane2019algorithmic}
\bibinfo{author}{A.~Birhane}, \bibinfo{author}{F.~Cummins},
\newblock \bibinfo{title}{Algorithmic injustices: Towards a relational ethics},
\newblock \bibinfo{journal}{arXiv preprint arXiv:1912.07376}
  (\bibinfo{year}{2019}).
\bibitem[{Verbeek(2008)}]{verbeek2008morality}
\bibinfo{author}{P.-P. Verbeek},
\newblock \bibinfo{title}{Morality in design: Design ethics and the morality of
  technological artifacts},
\newblock in: \bibinfo{booktitle}{Philosophy and design},
  \bibinfo{publisher}{Springer}, \bibinfo{year}{2008}, pp.
  \bibinfo{pages}{91--103}.
\bibitem[{Vayena et~al.(2018)Vayena, Blasimme, and Cohen}]{vayena2018machine}
\bibinfo{author}{E.~Vayena}, \bibinfo{author}{A.~Blasimme},
  \bibinfo{author}{I.~G. Cohen},
\newblock \bibinfo{title}{Machine learning in medicine: Addressing ethical
  challenges},
\newblock \bibinfo{journal}{PLoS medicine} \bibinfo{volume}{15}
  (\bibinfo{year}{2018}) \bibinfo{pages}{e1002689}.
\bibitem[{Hagendorff(2020)}]{hagendorff2020ethics}
\bibinfo{author}{T.~Hagendorff},
\newblock \bibinfo{title}{The ethics of ai ethics: An evaluation of
  guidelines},
\newblock \bibinfo{journal}{Minds and Machines}  (\bibinfo{year}{2020})
  \bibinfo{pages}{1--22}.
\bibitem[{Torresen(2018)}]{torresen2018review}
\bibinfo{author}{J.~Torresen},
\newblock \bibinfo{title}{A review of future and ethical perspectives of
  robotics and ai},
\newblock \bibinfo{journal}{Frontiers in Robotics and AI} \bibinfo{volume}{4}
  (\bibinfo{year}{2018}) \bibinfo{pages}{75}.
\bibitem[{Lin et~al.(2012)Lin, Abney, and Bekey}]{lin2012robot}
\bibinfo{author}{P.~Lin}, \bibinfo{author}{K.~Abney}, \bibinfo{author}{G.~A.
  Bekey}, \bibinfo{title}{Robot ethics: the ethical and social implications of
  robotics}, \bibinfo{publisher}{Intelligent Robotics and Autonomous Agents
  series}, \bibinfo{year}{2012}.
\bibitem[{Anderson and Anderson(2010)}]{anderson2010robot}
\bibinfo{author}{M.~Anderson}, \bibinfo{author}{S.~L. Anderson},
\newblock \bibinfo{title}{Robot be good},
\newblock \bibinfo{journal}{Scientific American} \bibinfo{volume}{303}
  (\bibinfo{year}{2010}) \bibinfo{pages}{72--77}.
\bibitem[{Boden et~al.(2017)Boden, Bryson, Caldwell, Dautenhahn, Edwards,
  Kember, Newman, Parry, Pegman, Rodden et~al.}]{boden2017principles}
\bibinfo{author}{M.~Boden}, \bibinfo{author}{J.~Bryson},
  \bibinfo{author}{D.~Caldwell}, \bibinfo{author}{K.~Dautenhahn},
  \bibinfo{author}{L.~Edwards}, \bibinfo{author}{S.~Kember},
  \bibinfo{author}{P.~Newman}, \bibinfo{author}{V.~Parry},
  \bibinfo{author}{G.~Pegman}, \bibinfo{author}{T.~Rodden}, et~al.,
\newblock \bibinfo{title}{Principles of robotics: regulating robots in the real
  world},
\newblock \bibinfo{journal}{Connection Science} \bibinfo{volume}{29}
  (\bibinfo{year}{2017}) \bibinfo{pages}{124--129}.
\bibitem[{BSI-2016(2016)}]{bsi2016bs}
\bibinfo{author}{BSI-2016},
\newblock \bibinfo{title}{Bs 8611: 2016 robots and robotic devices: Guide to
  the ethical design and application of robots and robotic systems},
\newblock \bibinfo{journal}{British Standards Institution}
  (\bibinfo{year}{2016}).
\bibitem[{Reed et~al.(2016)Reed, Kennedy, and Silva}]{reed2016responsibility}
\bibinfo{author}{C.~Reed}, \bibinfo{author}{E.~Kennedy},
  \bibinfo{author}{S.~Silva},
\newblock \bibinfo{title}{Responsibility, autonomy and accountability: legal
  liability for machine learning},
\newblock \bibinfo{journal}{Queen Mary School of Law Legal Studies Research
  Paper}  (\bibinfo{year}{2016}).
\bibitem[{Johnson et~al.(2019)Johnson, Pasquale, and
  Chapman}]{johnson2019artificial}
\bibinfo{author}{K.~Johnson}, \bibinfo{author}{F.~Pasquale},
  \bibinfo{author}{J.~Chapman},
\newblock \bibinfo{title}{Artificial intelligence, machine learning, and bias
  in finance: toward responsible innovation},
\newblock \bibinfo{journal}{Fordham L. Rev.} \bibinfo{volume}{88}
  (\bibinfo{year}{2019}) \bibinfo{pages}{499}.
\bibitem[{Arrieta et~al.(2020)Arrieta, D{\'\i}az-Rodr{\'\i}guez, Del~Ser,
  Bennetot, Tabik, Barbado, Garc{\'\i}a, Gil-L{\'o}pez, Molina, Benjamins
  et~al.}]{arrieta2020explainable}
\bibinfo{author}{A.~B. Arrieta}, \bibinfo{author}{N.~D{\'\i}az-Rodr{\'\i}guez},
  \bibinfo{author}{J.~Del~Ser}, \bibinfo{author}{A.~Bennetot},
  \bibinfo{author}{S.~Tabik}, \bibinfo{author}{A.~Barbado},
  \bibinfo{author}{S.~Garc{\'\i}a}, \bibinfo{author}{S.~Gil-L{\'o}pez},
  \bibinfo{author}{D.~Molina}, \bibinfo{author}{R.~Benjamins}, et~al.,
\newblock \bibinfo{title}{Explainable artificial intelligence (xai): Concepts,
  taxonomies, opportunities and challenges toward responsible ai},
\newblock \bibinfo{journal}{Information Fusion} \bibinfo{volume}{58}
  (\bibinfo{year}{2020}) \bibinfo{pages}{82--115}.
\bibitem[{Goodman and Flaxman(2017)}]{goodman2017european}
\bibinfo{author}{B.~Goodman}, \bibinfo{author}{S.~Flaxman},
\newblock \bibinfo{title}{European union regulations on algorithmic
  decision-making and a “right to explanation”},
\newblock \bibinfo{journal}{AI magazine} \bibinfo{volume}{38}
  (\bibinfo{year}{2017}) \bibinfo{pages}{50--57}.
\bibitem[{De~Santis et~al.(2008)De~Santis, Siciliano, De~Luca, and
  Bicchi}]{de2008atlas}
\bibinfo{author}{A.~De~Santis}, \bibinfo{author}{B.~Siciliano},
  \bibinfo{author}{A.~De~Luca}, \bibinfo{author}{A.~Bicchi},
\newblock \bibinfo{title}{An atlas of physical human--robot interaction},
\newblock \bibinfo{journal}{Mechanism and Machine Theory} \bibinfo{volume}{43}
  (\bibinfo{year}{2008}) \bibinfo{pages}{253--270}.
\bibitem[{Vandemeulebroucke et~al.(2020)Vandemeulebroucke, de~Casterl{\'e}, and
  Gastmans}]{vandemeulebroucke2020ethics}
\bibinfo{author}{T.~Vandemeulebroucke}, \bibinfo{author}{B.~D.
  de~Casterl{\'e}}, \bibinfo{author}{C.~Gastmans},
\newblock \bibinfo{title}{Ethics of socially assistive robots in aged-care
  settings: a socio-historical contextualisation},
\newblock \bibinfo{journal}{Journal of Medical Ethics} \bibinfo{volume}{46}
  (\bibinfo{year}{2020}) \bibinfo{pages}{128--136}.
\bibitem[{Riek and Howard(2014)}]{riek2014code}
\bibinfo{author}{L.~Riek}, \bibinfo{author}{D.~Howard},
\newblock \bibinfo{title}{A code of ethics for the human-robot interaction
  profession},
\newblock \bibinfo{journal}{Proceedings of We Robot}  (\bibinfo{year}{2014}).
\bibitem[{Bonnefon et~al.(2020)Bonnefon, {\v{C}}erny, Danaher, Devillier,
  Johansson, Kovacikova, Martens, Mladenovic, Palade, Reed
  et~al.}]{bonnefon2020ethics}
\bibinfo{author}{J.-F. Bonnefon}, \bibinfo{author}{D.~{\v{C}}erny},
  \bibinfo{author}{J.~Danaher}, \bibinfo{author}{N.~Devillier},
  \bibinfo{author}{V.~Johansson}, \bibinfo{author}{T.~Kovacikova},
  \bibinfo{author}{M.~Martens}, \bibinfo{author}{M.~Mladenovic},
  \bibinfo{author}{P.~Palade}, \bibinfo{author}{N.~Reed}, et~al.,
\newblock \bibinfo{title}{Ethics of connected and automated vehicles:
  Recommendations on road safety, privacy, fairness, explainability and
  responsibility},
\newblock \bibinfo{journal}{EU publications}  (\bibinfo{year}{2020}).
\bibitem[{Brand{\~a}o et~al.(2020)Brand{\~a}o, Jirtoka, Webb, and
  Luff}]{brandao2020fair}
\bibinfo{author}{M.~Brand{\~a}o}, \bibinfo{author}{M.~Jirtoka},
  \bibinfo{author}{H.~Webb}, \bibinfo{author}{P.~Luff},
\newblock \bibinfo{title}{Fair navigation planning: A resource for
  characterizing and designing fairness in mobile robots},
\newblock \bibinfo{journal}{Artificial Intelligence}  (\bibinfo{year}{2020})
  \bibinfo{pages}{103259}.
\bibitem[{Castro and Toro(2004)}]{castro2004evolution}
\bibinfo{author}{L.~Castro}, \bibinfo{author}{M.~A. Toro},
\newblock \bibinfo{title}{The evolution of culture: from primate social
  learning to human culture},
\newblock \bibinfo{journal}{Proceedings of the National Academy of Sciences}
  \bibinfo{volume}{101} (\bibinfo{year}{2004}) \bibinfo{pages}{10235--10240}.
\bibitem[{Kaushal et~al.(2020)Kaushal, Altman, and Langlotz}]{amit2020}
\bibinfo{author}{A.~Kaushal}, \bibinfo{author}{R.~Altman},
  \bibinfo{author}{C.~Langlotz},
\newblock \bibinfo{title}{Health care ai systems are biased},
\newblock \bibinfo{journal}{SCIENTIFIC AMERICAN}  (\bibinfo{year}{2020}).
  \URLprefix
  \url{https://www.scientificamerican.com/article/health-care-ai-systems-are-biased/}.
\bibitem[{Thrun et~al.(2000)Thrun, Schulte, and
  Rosenberg}]{thrun2000interaction}
\bibinfo{author}{S.~Thrun}, \bibinfo{author}{J.~Schulte},
  \bibinfo{author}{C.~Rosenberg},
\newblock \bibinfo{title}{Interaction with mobile robots in public places},
\newblock \bibinfo{journal}{IEEE Intelligent Systems}  (\bibinfo{year}{2000})
  \bibinfo{pages}{7--11}.
\bibitem[{Yu(2019)}]{yu2019direct}
\bibinfo{author}{A.~Yu},
\newblock \bibinfo{title}{Direct discrimination and indirect discrimination: A
  distinction with a difference},
\newblock \bibinfo{journal}{WJ Legal Stud.} \bibinfo{volume}{9}
  (\bibinfo{year}{2019}) \bibinfo{pages}{1}.
\bibitem[{Forshaw and Pilgerstorfer(2008)}]{forshaw2008direct}
\bibinfo{author}{S.~Forshaw}, \bibinfo{author}{M.~Pilgerstorfer},
\newblock \bibinfo{title}{Direct and indirect discrimination: is there
  something in between?},
\newblock \bibinfo{journal}{Industrial Law Journal} \bibinfo{volume}{37}
  (\bibinfo{year}{2008}) \bibinfo{pages}{347--364}.
\bibitem[{Zhang et~al.(2016)Zhang, Wu, and Wu}]{zhang2016causal}
\bibinfo{author}{L.~Zhang}, \bibinfo{author}{Y.~Wu}, \bibinfo{author}{X.~Wu},
\newblock \bibinfo{title}{A causal framework for discovering and removing
  direct and indirect discrimination},
\newblock \bibinfo{journal}{arXiv preprint arXiv:1611.07509}
  (\bibinfo{year}{2016}).
\bibitem[{{\"O}tting et~al.(2017){\"O}tting, Gopinathan, Maier, and
  Steil}]{otting2017criteria}
\bibinfo{author}{S.~K. {\"O}tting}, \bibinfo{author}{S.~Gopinathan},
  \bibinfo{author}{G.~W. Maier}, \bibinfo{author}{J.~J. Steil},
\newblock \bibinfo{title}{Why criteria of decision fairness should be
  considered in robot design},
\newblock \bibinfo{journal}{The 20th ACM Conference on Computer-Supported
  Cooperative Work and Social Computing}  (\bibinfo{year}{2017}).
\bibitem[{Claure et~al.(2019)Claure, Chen, Modi, Jung, and
  Nikolaidis}]{claure2019reinforcement}
\bibinfo{author}{H.~Claure}, \bibinfo{author}{Y.~Chen},
  \bibinfo{author}{J.~Modi}, \bibinfo{author}{M.~Jung},
  \bibinfo{author}{S.~Nikolaidis},
\newblock \bibinfo{title}{Reinforcement learning with fairness constraints for
  resource distribution in human-robot teams},
\newblock \bibinfo{journal}{arXiv preprint arXiv:1907.00313}
  (\bibinfo{year}{2019}).
\bibitem[{Lee(2018)}]{lee2018detecting}
\bibinfo{author}{N.~T. Lee},
\newblock \bibinfo{title}{Detecting racial bias in algorithms and machine
  learning},
\newblock \bibinfo{journal}{Journal of Information, Communication and Ethics in
  Society}  (\bibinfo{year}{2018}).
\bibitem[{Nelson(2019)}]{nelson2019bias}
\bibinfo{author}{G.~S. Nelson},
\newblock \bibinfo{title}{Bias in artificial intelligence},
\newblock \bibinfo{journal}{North Carolina medical journal}
  \bibinfo{volume}{80} (\bibinfo{year}{2019}) \bibinfo{pages}{220--222}.
\bibitem[{Fuchs(2018)}]{fuchs2018dangers}
\bibinfo{author}{D.~J. Fuchs},
\newblock \bibinfo{title}{The dangers of human-like bias in machine-learning
  algorithms},
\newblock \bibinfo{journal}{Missouri S\&T’s Peer to Peer} \bibinfo{volume}{2}
  (\bibinfo{year}{2018}) \bibinfo{pages}{1}.
\bibitem[{Howard et~al.(2017)Howard, Zhang, and Horvitz}]{howard2017addressing}
\bibinfo{author}{A.~Howard}, \bibinfo{author}{C.~Zhang},
  \bibinfo{author}{E.~Horvitz},
\newblock \bibinfo{title}{Addressing bias in machine learning algorithms: A
  pilot study on emotion recognition for intelligent systems},
\newblock in: \bibinfo{booktitle}{2017 IEEE Workshop on Advanced Robotics and
  its Social Impacts (ARSO)}, \bibinfo{organization}{IEEE},
  \bibinfo{year}{2017}, pp. \bibinfo{pages}{1--7}.
\bibitem[{Chouldechova and Roth(2018)}]{chouldechova2018frontiers}
\bibinfo{author}{A.~Chouldechova}, \bibinfo{author}{A.~Roth},
\newblock \bibinfo{title}{The frontiers of fairness in machine learning},
\newblock \bibinfo{journal}{arXiv preprint arXiv:1810.08810}
  (\bibinfo{year}{2018}).
\bibitem[{Binns(2018)}]{binns2018fairness}
\bibinfo{author}{R.~Binns},
\newblock \bibinfo{title}{Fairness in machine learning: Lessons from political
  philosophy},
\newblock in: \bibinfo{booktitle}{Conference on Fairness, Accountability and
  Transparency}, \bibinfo{organization}{PMLR}, \bibinfo{year}{2018}, pp.
  \bibinfo{pages}{149--159}.
\bibitem[{Winner(1978)}]{winner1978autonomous}
\bibinfo{author}{L.~Winner}, \bibinfo{title}{Autonomous technology:
  Technics-out-of-control as a theme in political thought},
  \bibinfo{publisher}{Mit Press}, \bibinfo{year}{1978}.
\bibitem[{Grunwald(2011)}]{grunwald2011responsible}
\bibinfo{author}{A.~Grunwald},
\newblock \bibinfo{title}{Responsible innovation: bringing together technology
  assessment, applied ethics, and sts research},
\newblock \bibinfo{journal}{Enterprise and Work Innovation Studies}
  \bibinfo{volume}{31} (\bibinfo{year}{2011}) \bibinfo{pages}{10--9}.
\bibitem[{Stilgoe et~al.(2013)Stilgoe, Owen, and
  Macnaghten}]{stilgoe2013developing}
\bibinfo{author}{J.~Stilgoe}, \bibinfo{author}{R.~Owen},
  \bibinfo{author}{P.~Macnaghten},
\newblock \bibinfo{title}{Developing a framework for responsible innovation},
\newblock \bibinfo{journal}{Research policy} \bibinfo{volume}{42}
  (\bibinfo{year}{2013}) \bibinfo{pages}{1568--1580}.
\bibitem[{Simmel(1949)}]{simmel1949sociology}
\bibinfo{author}{G.~Simmel},
\newblock \bibinfo{title}{The sociology of sociability},
\newblock \bibinfo{journal}{American journal of sociology} \bibinfo{volume}{55}
  (\bibinfo{year}{1949}) \bibinfo{pages}{254--261}.
\bibitem[{Fehr and Fischbacher(2004)}]{fehr2004social}
\bibinfo{author}{E.~Fehr}, \bibinfo{author}{U.~Fischbacher},
\newblock \bibinfo{title}{Social norms and human cooperation},
\newblock \bibinfo{journal}{Trends in cognitive sciences} \bibinfo{volume}{8}
  (\bibinfo{year}{2004}) \bibinfo{pages}{185--190}.
\bibitem[{Kirby(2010)}]{kirby2010social}
\bibinfo{author}{R.~Kirby}, \bibinfo{title}{Social Robot Navigation}, Ph.D.
  thesis, Carnegie Mellon University, \bibinfo{address}{Pittsburgh, PA},
  \bibinfo{year}{2010}.
\bibitem[{Hall et~al.(1968)Hall, Birdwhistell, Bock, Bohannan, Diebold~Jr,
  Durbin, Edmonson, Fischer, Hymes, Kimball et~al.}]{hall1968proxemics}
\bibinfo{author}{E.~T. Hall}, \bibinfo{author}{R.~L. Birdwhistell},
  \bibinfo{author}{B.~Bock}, \bibinfo{author}{P.~Bohannan},
  \bibinfo{author}{A.~R. Diebold~Jr}, \bibinfo{author}{M.~Durbin},
  \bibinfo{author}{M.~S. Edmonson}, \bibinfo{author}{J.~Fischer},
  \bibinfo{author}{D.~Hymes}, \bibinfo{author}{S.~T. Kimball}, et~al.,
\newblock \bibinfo{title}{Proxemics [and comments and replies]},
\newblock \bibinfo{journal}{Current anthropology} \bibinfo{volume}{9}
  (\bibinfo{year}{1968}) \bibinfo{pages}{83--108}.
\bibitem[{Birdwhistell(2010)}]{birdwhistell2010kinesics}
\bibinfo{author}{R.~L. Birdwhistell}, \bibinfo{title}{Kinesics and context:
  Essays on body motion communication}, \bibinfo{publisher}{University of
  Pennsylvania press}, \bibinfo{year}{2010}.
\bibitem[{Argyle et~al.(1994)Argyle, Cook, and Cramer}]{argyle1976gaze}
\bibinfo{author}{M.~Argyle}, \bibinfo{author}{M.~Cook},
  \bibinfo{author}{D.~Cramer},
\newblock \bibinfo{title}{Gaze and mutual gaze},
\newblock \bibinfo{journal}{British Journal of Psychiatry}
  \bibinfo{volume}{165} (\bibinfo{year}{1994}) \bibinfo{pages}{848–850}.
\bibitem[{Rios-Martinez et~al.(2015)Rios-Martinez, Spalanzani, and
  Laugier}]{rios2015proxemics}
\bibinfo{author}{J.~Rios-Martinez}, \bibinfo{author}{A.~Spalanzani},
  \bibinfo{author}{C.~Laugier},
\newblock \bibinfo{title}{From proxemics theory to socially-aware navigation: A
  survey},
\newblock \bibinfo{journal}{International Journal of Social Robotics}
  \bibinfo{volume}{7} (\bibinfo{year}{2015}) \bibinfo{pages}{137--153}.
\bibitem[{Birdwhistell(1952)}]{birdwhistell1952introduction}
\bibinfo{author}{R.~L. Birdwhistell}, \bibinfo{title}{Introduction to kinesics:
  An annotation system for analysis of body motion and gesture},
  \bibinfo{publisher}{Department of State, Foreign Service Institute},
  \bibinfo{year}{1952}.
\bibitem[{Harrigan(2005)}]{harrigan2005proxemics}
\bibinfo{author}{J.~A. Harrigan}, \bibinfo{title}{Proxemics, kinesics, and
  gaze}, \bibinfo{publisher}{Oxford University Press}, \bibinfo{year}{2005}.
\bibitem[{Fiorini and Prassler(2000)}]{fiorini2000cleaning}
\bibinfo{author}{P.~Fiorini}, \bibinfo{author}{E.~Prassler},
\newblock \bibinfo{title}{Cleaning and household robots: A technology survey},
\newblock \bibinfo{journal}{Autonomous robots} \bibinfo{volume}{9}
  (\bibinfo{year}{2000}) \bibinfo{pages}{227--235}.
\bibitem[{Helbing and Molnar(1995)}]{helbing1995social}
\bibinfo{author}{D.~Helbing}, \bibinfo{author}{P.~Molnar},
\newblock \bibinfo{title}{Social force model for pedestrian dynamics},
\newblock \bibinfo{journal}{Physical review E} \bibinfo{volume}{51}
  (\bibinfo{year}{1995}) \bibinfo{pages}{4282}.
\bibitem[{Ferrer et~al.(2017)Ferrer, Zulueta, Cotarelo, and
  Sanfeliu}]{ferrer2017robot}
\bibinfo{author}{G.~Ferrer}, \bibinfo{author}{A.~G. Zulueta},
  \bibinfo{author}{F.~H. Cotarelo}, \bibinfo{author}{A.~Sanfeliu},
\newblock \bibinfo{title}{Robot social-aware navigation framework to accompany
  people walking side-by-side},
\newblock \bibinfo{journal}{Autonomous robots} \bibinfo{volume}{41}
  (\bibinfo{year}{2017}) \bibinfo{pages}{775--793}.
\bibitem[{Tai et~al.(2018)Tai, Zhang, Liu, and Burgard}]{tai2018socially}
\bibinfo{author}{L.~Tai}, \bibinfo{author}{J.~Zhang}, \bibinfo{author}{M.~Liu},
  \bibinfo{author}{W.~Burgard},
\newblock \bibinfo{title}{Socially compliant navigation through raw depth
  inputs with generative adversarial imitation learning},
\newblock in: \bibinfo{booktitle}{2018 IEEE International Conference on
  Robotics and Automation (ICRA)}, \bibinfo{organization}{IEEE},
  \bibinfo{year}{2018}, pp. \bibinfo{pages}{1111--1117}.
\bibitem[{Groshev et~al.(2017)Groshev, Goldstein, Tamar, Srivastava, and
  Abbeel}]{groshev2017learning}
\bibinfo{author}{E.~Groshev}, \bibinfo{author}{M.~Goldstein},
  \bibinfo{author}{A.~Tamar}, \bibinfo{author}{S.~Srivastava},
  \bibinfo{author}{P.~Abbeel},
\newblock \bibinfo{title}{Learning generalized reactive policies using deep
  neural networks},
\newblock \bibinfo{journal}{arXiv preprint arXiv:1708.07280}
  (\bibinfo{year}{2017}).
\bibitem[{Chen et~al.(2017)Chen, Everett, Liu, and How}]{chen2017socially}
\bibinfo{author}{Y.~F. Chen}, \bibinfo{author}{M.~Everett},
  \bibinfo{author}{M.~Liu}, \bibinfo{author}{J.~P. How},
\newblock \bibinfo{title}{Socially aware motion planning with deep
  reinforcement learning},
\newblock in: \bibinfo{booktitle}{2017 IEEE/RSJ International Conference on
  Intelligent Robots and Systems (IROS)}, \bibinfo{organization}{IEEE},
  \bibinfo{year}{2017}, pp. \bibinfo{pages}{1343--1350}.
\bibitem[{Kuderer et~al.(2013)Kuderer, Kretzschmar, and
  Burgard}]{kuderer2013teaching}
\bibinfo{author}{M.~Kuderer}, \bibinfo{author}{H.~Kretzschmar},
  \bibinfo{author}{W.~Burgard},
\newblock \bibinfo{title}{Teaching mobile robots to cooperatively navigate in
  populated environments},
\newblock in: \bibinfo{booktitle}{2013 IEEE/RSJ International Conference on
  Intelligent Robots and Systems}, \bibinfo{organization}{IEEE},
  \bibinfo{year}{2013}, pp. \bibinfo{pages}{3138--3143}.
\bibitem[{Hamandi et~al.(2019)Hamandi, D’Arcy, and
  Fazli}]{hamandi2019deepmotion}
\bibinfo{author}{M.~Hamandi}, \bibinfo{author}{M.~D’Arcy},
  \bibinfo{author}{P.~Fazli},
\newblock \bibinfo{title}{Deepmotion: Learning to navigate like humans},
\newblock in: \bibinfo{booktitle}{2019 28th IEEE International Conference on
  Robot and Human Interactive Communication (RO-MAN)},
  \bibinfo{organization}{IEEE}, \bibinfo{year}{2019}, pp.
  \bibinfo{pages}{1--7}.
\bibitem[{Watkins and Dayan(1992)}]{watkins1992q}
\bibinfo{author}{C.~J. Watkins}, \bibinfo{author}{P.~Dayan},
\newblock \bibinfo{title}{Q-learning},
\newblock \bibinfo{journal}{Machine learning} \bibinfo{volume}{8}
  (\bibinfo{year}{1992}) \bibinfo{pages}{279--292}.
\bibitem[{Kalweit et~al.(2020{\natexlab{a}})Kalweit, Huegle, Werling, and
  Boedecker}]{kalweit2020deep}
\bibinfo{author}{G.~Kalweit}, \bibinfo{author}{M.~Huegle},
  \bibinfo{author}{M.~Werling}, \bibinfo{author}{J.~Boedecker},
\newblock \bibinfo{title}{Deep inverse q-learning with constraints},
\newblock \bibinfo{journal}{Advances in Neural Information Processing Systems}
  \bibinfo{volume}{33} (\bibinfo{year}{2020}{\natexlab{a}}).
\bibitem[{Kalweit et~al.(2020{\natexlab{b}})Kalweit, Huegle, Werling, and
  Boedecker}]{kalweit2020interpretable}
\bibinfo{author}{G.~Kalweit}, \bibinfo{author}{M.~Huegle},
  \bibinfo{author}{M.~Werling}, \bibinfo{author}{J.~Boedecker},
\newblock \bibinfo{title}{Interpretable multi time-scale constraints in
  model-free deep reinforcement learning for autonomous driving},
\newblock \bibinfo{journal}{arXiv preprint arXiv:2003.09398}
  (\bibinfo{year}{2020}{\natexlab{b}}).
\bibitem[{Geirhos et~al.(2020)Geirhos, Jacobsen, Michaelis, Zemel, Brendel,
  Bethge, and Wichmann}]{geirhos2020shortcut}
\bibinfo{author}{R.~Geirhos}, \bibinfo{author}{J.-H. Jacobsen},
  \bibinfo{author}{C.~Michaelis}, \bibinfo{author}{R.~Zemel},
  \bibinfo{author}{W.~Brendel}, \bibinfo{author}{M.~Bethge},
  \bibinfo{author}{F.~A. Wichmann},
\newblock \bibinfo{title}{Shortcut learning in deep neural networks},
\newblock \bibinfo{journal}{arXiv preprint arXiv:2004.07780}
  (\bibinfo{year}{2020}).
\bibitem[{Dastin(2018)}]{dastin2018amazon}
\bibinfo{author}{J.~Dastin},
\newblock \bibinfo{title}{Amazon scraps secret ai recruiting tool that showed
  bias against women},
\newblock \bibinfo{journal}{San Fransico, CA: Reuters. Retrieved on October}
  \bibinfo{volume}{9} (\bibinfo{year}{2018}) \bibinfo{pages}{2018}.
\bibitem[{Patompak et~al.(2019)Patompak, Jeong, Nilkhamhang, and
  Chong}]{patompak2019learning}
\bibinfo{author}{P.~Patompak}, \bibinfo{author}{S.~Jeong},
  \bibinfo{author}{I.~Nilkhamhang}, \bibinfo{author}{N.~Y. Chong},
\newblock \bibinfo{title}{Learning proxemics for personalized human--robot
  social interaction},
\newblock \bibinfo{journal}{International Journal of Social Robotics}
  (\bibinfo{year}{2019}) \bibinfo{pages}{1--14}.
\bibitem[{Kivrak et~al.(2020)Kivrak, Cakmak, Kose, and
  Yavuz}]{kivrak2020social}
\bibinfo{author}{H.~Kivrak}, \bibinfo{author}{F.~Cakmak},
  \bibinfo{author}{H.~Kose}, \bibinfo{author}{S.~Yavuz},
\newblock \bibinfo{title}{Social navigation framework for assistive robots in
  human inhabited unknown environments},
\newblock \bibinfo{journal}{Engineering Science and Technology, an
  International Journal}  (\bibinfo{year}{2020}).
\bibitem[{Jones and Healy(2006)}]{jones2006differences}
\bibinfo{author}{C.~M. Jones}, \bibinfo{author}{S.~D. Healy},
\newblock \bibinfo{title}{Differences in cue use and spatial memory in men and
  women},
\newblock \bibinfo{journal}{Proceedings of the Royal Society B: Biological
  Sciences} \bibinfo{volume}{273} (\bibinfo{year}{2006})
  \bibinfo{pages}{2241--2247}.
\bibitem[{S{\"o}derstr{\"o}m(2001)}]{soderstrom2001researchers}
\bibinfo{author}{M.~S{\"o}derstr{\"o}m},
\newblock \bibinfo{title}{Why researchers excluded women from their trial
  populations},
\newblock \bibinfo{journal}{Lakartidningen} \bibinfo{volume}{98}
  (\bibinfo{year}{2001}) \bibinfo{pages}{1524--1528}.
\bibitem[{Perez(2016)}]{perez2016microsoft}
\bibinfo{author}{S.~Perez},
\newblock \bibinfo{title}{Microsoft silences its new ai bot tay, after twitter
  users teach it racism},
\newblock \bibinfo{journal}{Tech Crunch}  (\bibinfo{year}{2016}).
\bibitem[{Aljalbout et~al.(2018)Aljalbout, Golkov, Siddiqui, Strobel, and
  Cremers}]{aljalbout2018clustering}
\bibinfo{author}{E.~Aljalbout}, \bibinfo{author}{V.~Golkov},
  \bibinfo{author}{Y.~Siddiqui}, \bibinfo{author}{M.~Strobel},
  \bibinfo{author}{D.~Cremers},
\newblock \bibinfo{title}{Clustering with deep learning: Taxonomy and new
  methods},
\newblock \bibinfo{journal}{arXiv preprint arXiv:1801.07648}
  (\bibinfo{year}{2018}).
\bibitem[{Icograms(2020)}]{icograms}
\bibinfo{author}{Icograms}, \bibinfo{title}{Illustrations},
  \bibinfo{howpublished}{\url{https://icograms.com/}}, \bibinfo{year}{2020}.
  \bibinfo{note}{Accessed: 2020-12-18}.
\bibitem[{Fink et~al.(2013)Fink, Bauwens, Kaplan, and
  Dillenbourg}]{fink2013living}
\bibinfo{author}{J.~Fink}, \bibinfo{author}{V.~Bauwens},
  \bibinfo{author}{F.~Kaplan}, \bibinfo{author}{P.~Dillenbourg},
\newblock \bibinfo{title}{Living with a vacuum cleaning robot},
\newblock \bibinfo{journal}{International Journal of Social Robotics}
  \bibinfo{volume}{5} (\bibinfo{year}{2013}) \bibinfo{pages}{389--408}.
\bibitem[{Forlizzi and DiSalvo(2006)}]{forlizzi2006service}
\bibinfo{author}{J.~Forlizzi}, \bibinfo{author}{C.~DiSalvo},
\newblock \bibinfo{title}{Service robots in the domestic environment: a study
  of the roomba vacuum in the home},
\newblock in: \bibinfo{booktitle}{Proceedings of the 1st ACM SIGCHI/SIGART
  conference on Human-robot interaction}, \bibinfo{year}{2006}, pp.
  \bibinfo{pages}{258--265}.
\bibitem[{Forlizzi(2007)}]{forlizzi2007robotic}
\bibinfo{author}{J.~Forlizzi},
\newblock \bibinfo{title}{How robotic products become social products: an
  ethnographic study of cleaning in the home},
\newblock in: \bibinfo{booktitle}{2007 2nd ACM/IEEE International Conference on
  Human-Robot Interaction (HRI)}, \bibinfo{organization}{IEEE},
  \bibinfo{year}{2007}, pp. \bibinfo{pages}{129--136}.
\bibitem[{Research(2019)}]{clearningMarket}
\bibinfo{author}{I.~Research},
\newblock \bibinfo{title}{Floor cleaning robot market by robot type , by sales
  channel, by region - global forecast up to 2025},
\newblock \bibinfo{journal}{Research and Markets}  (\bibinfo{year}{2019}).
  \URLprefix
  \url{https://www.researchandmarkets.com/reports/4866505/floor-cleaning-robot-market-by-robot-type-by?utm_source=dynamic&utm_medium=CI&utm_code=4px2dc&utm_campaign=1328922+-+Global+Floor+Cleaning+Robot+Market+2019-2025%3a+Increasing+Potential+for+AI+Empowered+Cleaning+Robots+in+Commercial+Applications+&utm_exec=chdo54cid}.
\bibitem[{Brandao(2019)}]{brandao2019age}
\bibinfo{author}{M.~Brandao},
\newblock \bibinfo{title}{Age and gender bias in pedestrian detection
  algorithms},
\newblock \bibinfo{journal}{arXiv preprint arXiv:1906.10490}
  (\bibinfo{year}{2019}).
\bibitem[{Prabhu and Birhane(2020)}]{prabhu2020large}
\bibinfo{author}{V.~U. Prabhu}, \bibinfo{author}{A.~Birhane},
\newblock \bibinfo{title}{Large image datasets: A pyrrhic win for computer
  vision?},
\newblock \bibinfo{journal}{arXiv preprint arXiv:2006.16923}
  (\bibinfo{year}{2020}).
\bibitem[{Nottingham et~al.(2018)Nottingham, Johnson, and
  Russell}]{nottingham2018effect}
\bibinfo{author}{Q.~J. Nottingham}, \bibinfo{author}{D.~M. Johnson},
  \bibinfo{author}{R.~S. Russell},
\newblock \bibinfo{title}{The effect of waiting time on patient perceptions of
  care quality},
\newblock \bibinfo{journal}{Quality Management Journal} \bibinfo{volume}{25}
  (\bibinfo{year}{2018}) \bibinfo{pages}{32--45}.

\end{thebibliography}

\end{document}